\documentclass[lettersize,journal]{IEEEtran}
\usepackage{amsmath,amsfonts}
\usepackage{algorithmic}
\usepackage{array}
\usepackage[caption=false,font=normalsize,labelfont=sf,textfont=sf]{subfig}
\usepackage{textcomp}
\usepackage{stfloats}
\usepackage{url}
\usepackage[numbers,sort&compress]{natbib}
\usepackage[colorlinks=true,linkcolor=blue,citecolor=blue,urlcolor=blue,]{hyperref}
\usepackage{verbatim}
\usepackage{graphicx}
\usepackage{caption}  
\usepackage{makecell}
\usepackage{multirow}
\def\BibTeX{{\rm B\kern-.05em{\sc i\kern-.025em b}\kern-.08em
    T\kern-.1667em\lower.7ex\hbox{E}\kern-.125emX}}
\usepackage{balance}
\usepackage{colortbl}
\usepackage{xcolor}
\usepackage{booktabs}

\begin{document}

\title{Dense Feature Interaction Network for Image Inpainting Localization}

\author{Ye Yao, Tingfeng Han, Shan Jia, Siwei Lyu, \IEEEmembership{Fellow, IEEE}
\thanks{This work was supported by the National Natural Science Foundation of China under Grant Number 62471167, and the Zhejiang Provincial Natural Science Foundation of China under Grant Number LY24F020017. The authors acknowledge the Supercomputing Center of Hangzhou Dianzi University for providing computing resources. \textit{(Corresponding author: Shan Jia)}}
\thanks{Ye Yao and Tingfeng Han are with the School of Cyberspace, Hangzhou Dianzi University, Hangzhou, Zhejiang 310018, China (e-mail: yaoye@hdu.edu.cn).} 
\thanks{Shan Jia and Siwei Lyu are with the Department of Computer Science and Engineering, University at Buffalo, State University of New York, NY, USA (e-mail: shanjia@buffalo.edu).

This paper has been accepted for publication in IEEE Transactions on Information Forensics \& Security.}
}
%\author{IEEE Publication Technology,~\IEEEmembership{Staff,~IEEE,}
        % <-this % stops a space
%\thanks{This paper was produced by the IEEE Publication Technology Group. They are in Piscataway, NJ.}% <-this % stops a space
%\thanks{Manuscript received April 19, 2021; revised August 16, 2021.}}



\maketitle

\begin{abstract}
Image inpainting, the process of filling in missing areas in an image, is a common image editing technique. Inpainting can be used to conceal or alter image contents in malicious manipulation of images, driving the need for research in image inpainting detection. Most existing methods use a basic encoder-decoder structure, which often results in a high number of false positives or misses the inpainted regions, especially when dealing with targets of varying semantics and scales. Additionally, the lack of an effective approach to capture boundary artifacts leads to less accurate edge localization. 
In this paper, we describe a new method for inpainting detection based on a Dense Feature Interaction Network (DeFI-Net). DeFI-Net uses a novel feature pyramid architecture to capture and amplify multi-scale representations across various stages, thereby improving the detection of image inpainting by better strengthening feature-level interactions. Additionally, the network can adaptively direct the lower-level features, which carry edge and shape information, to refine the localization of manipulated regions while integrating the higher-level semantic features. Using DeFI-Net, we develop a method combining complementary representations to accurately identify inpainted areas.  Evaluation on seven image inpainting datasets demonstrates the effectiveness of our approach, which achieves state-of-the-art performance in detecting inpainting across diverse models. Code and models are available at \href{https://github.com/Boombb/DeFI-Net_Inpainting}{https://github.com/Boombb/DeFI-Net\_Inpainting}.
\end{abstract}


\section{Introduction}
\label{sec:intro}
Image inpainting is a common image editing technique that restores damaged or missing regions with visually credible structures and textures from the surrounding image content.  Inpainting has applications such as photo restoration, obstacle removal, and augmented reality.
Traditional inpainting methods formulate it as an optimization problem~\cite{xiang2023deep}, where the goal is to find the most suitable information to fill in the gaps while maintaining overall coherence in the reconstructed image. It is mainly divided into two branches: pixel propagation based~\cite{bertalmio2001navier, drori2003fragment, levin2003learning},  and image patches matching based~\cite{criminisi2004region, ting2007image, barnes2009patchmatch, darabi2012image, ding2018image}. The rapid advancements of learning-based methods, especially the recent generative AI tools such as OpenAI DALL-E\footnote{\url{https://openai.com/dall-e-2}}, Stable Diffusion\footnote{\url{https://huggingface.co/runwayml/stable-diffusion-inpainting}}, and Adobe Firefly\footnote{\url{https://www.adobe.com/products/firefly.html}} have enabled users to create and perform image editing, including inpainting, by simply providing some text prompts. The generative AI backed inpainting method exacerbates the negative impacts of malicious use of image inpainting as a means of disinformation. A notable recent example is the case where nude images created with AI-powered image inpainting were spread for cyber targeting female students in a California middle school\footnote{\url{https://www.latimes.com/california/story/2024-02-26/beverly-hills-middle-school-is-the-latest-to-be-rocked-by-deepfake-scandal}}.

\begin{figure}[t]
\centering
\includegraphics[width=\linewidth]{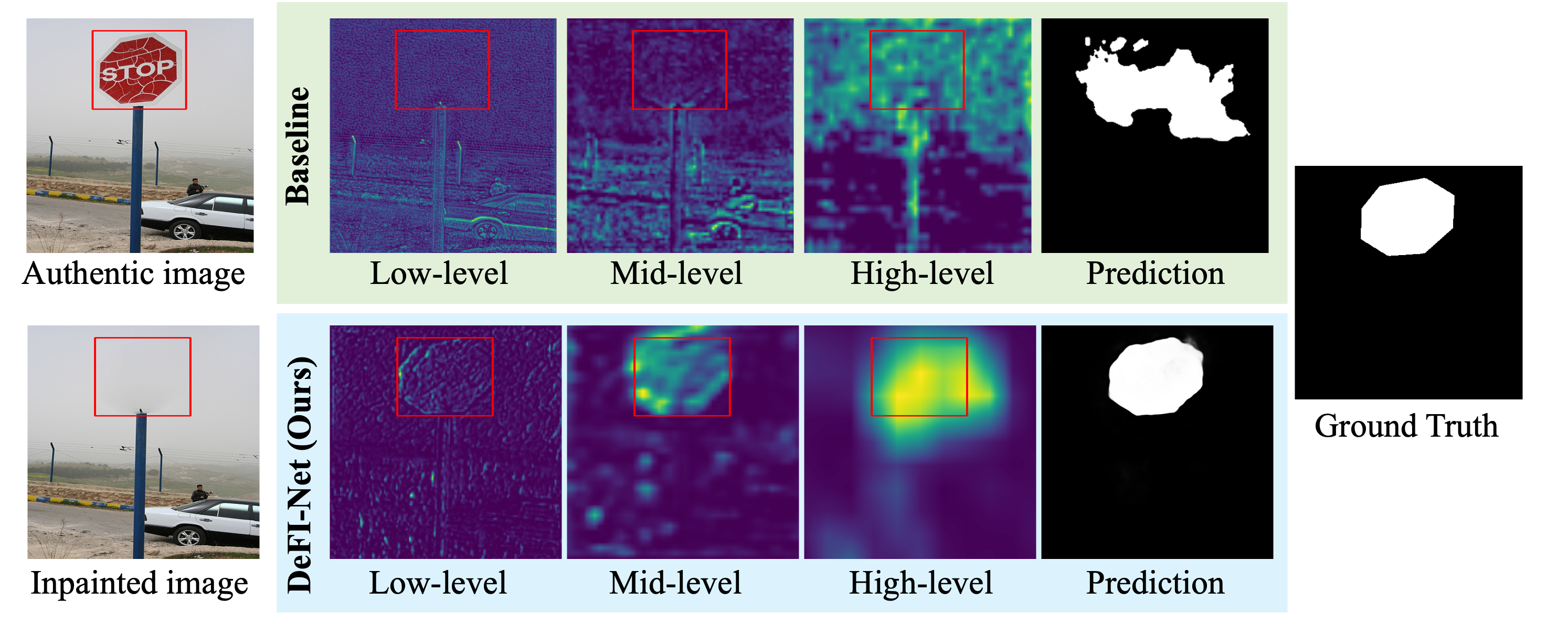}
\caption{Visualization of features at varying levels in describing inpainting clues. Yellow regions indicate higher responsiveness, highlighting key regions for identifying inpainting, while blue regions represent low responsiveness. Upper feature maps illustrate a baseline network PSCC-Net~\cite{liu2022pscc} that struggles to capture discriminative patterns despite employing multi-level feature learning. In contrast, the lower maps show our DeFI-Net, leveraging dense feature interaction to reveal diverse and complementary indicators crucial for accurate inpainting detection.}
\label{fig:teaser}
\end{figure}

To date, many methods have been developed to expose local image manipulation \cite{zhai2023towards, guo2023hierarchical, liu2022pscc, chen2021image, dong2022mvss, yu2024diffforensics}. Yet locating inpainted regions is still a challenging problem, which can be attributed to the following two factors. 
First, unlike traditional copy-move and splicing manipulations, which often produce noticeable artifacts such as duplicated pixels or inconsistent patterns in semantic and color, advanced image inpainting algorithms mitigate signal and visual artifacts by preserving statistical properties~\cite{chu2023rethinking} from both surrounding pixels and large-scale datasets in a data-driven manner. Additionally, the varied sizes of inpainted regions, ranging from large object editing to small watermark, signature, or trademark removal, pose further challenges in the detection process.
%tiny target inpainting detection is poor. And in real life, many malicious removal objects are usually smaller objects, such as watermarks, signatures or trademarks.

%First, image inpainting often involve complex and well-designed algorithms\cite{zhang2023image}, typically using deep learning models, to effectively mimic the surrounding image content and generate visually convincing regions. Second, unlike traditional copy-move and splicing manipulation, which leave distinct artifacts such as duplicated pixels\cite{li2018fast} or inconsistent compression clues~\cite{kwon2021cat} in the image, advanced image inpainting models can leave fewer signal artifacts by preserving the statistical properties \cite{chu2023rethinking} of the surrounding pixels. Furthermore, the inpainted region can have different sizes, further complicating the detection process. 

Although several studies have been dedicated exclusively to identifying image inpainting, such as using high-pass fully convolutional network~\cite{li2019localization}, Neural Architecture Search (NAS) based model~\cite{wu2021iid}, Transformer architecture~\cite{li2023transformer}, or primary–secondary network~\cite{zhang2023ps}, most adopt conventional encoder-decoder architectures. These methods prioritize high-level\footnote{
In this paper, we use ``level'' to indicate the abstraction level of extracted information, ``stage'' to describe a resolution convolution stream, as defined in HRNet~\cite{wang2020deep}, and ``layer'' as the basic building blocks of CNN architecture.} abstract information while overlooking the features that contain crucial contextual details, such as subtle inconsistencies in edges and shapes, which are essential for uncovering inpainting artifacts.
Furthermore, for learning inconsistencies at inpainting boundaries, relying solely on ordinary convolution combinations and fixed edge extraction operators makes it challenging to capture representative artifact features. 
Such limitations reduce the model's ability to address complex or small inpainting areas, resulting in false positives or diminished localization accuracy.

\begin{figure*}[t]
\centering
\includegraphics[width=0.9\linewidth]{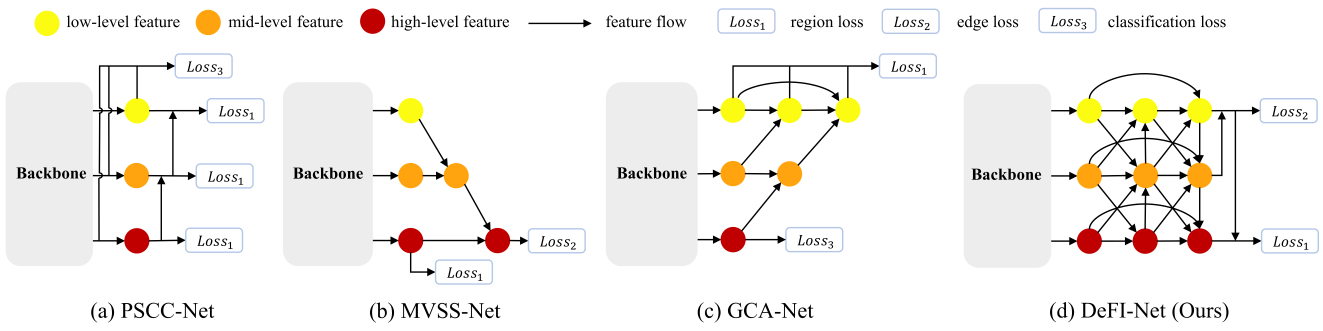}
\caption{A simplified schematic comparison between existing multi-scale learning strategies and our DeFI-Net. Unlike previous methods that rely on progressive interaction (e.g., PSCC-Net \cite{liu2022pscc}), focus more on high-level feature fusion (e.g., MVSS-Net \cite{dong2022mvss}), or low-level features (e.g., GCA-Net \cite{das2022gca}), our model offers dense feature interaction and adaptively utilizes complementary features from all levels to enhance inpainting localization. The yellow, orange, and red circles represent low-, mid-, and high-level features, respectively. The black arrows indicate the flow of features through the model. Finally, depending on the model design, the output is directed to different objective functions.}%The yellow, orange, and red circles represent low, mid, and high-level features, respectively. The black arrows indicate the flow of features. Finally, based on the model design, the input is directed to different objective functions.}}
\label{fig:module_compare}
\end{figure*}

We use the insight that features at various layers~\cite{lin2017feature} within deep learning networks can expose the subtle inconsistencies introduced by the inpainting operation in a different while complementary manner (\textit{refer to Fig.~\ref{fig:teaser}}). Low-level features with high resolution have smaller receptive fields, thus primarily capturing signal artifacts, like edges and {corners} lacking semantic information~\cite{lee2009convolutional}. Features at the middle level represent more abstract patterns and object parts, while high-level semantic features describing entire objects, scenes, or concepts are derived from the network's deeper layers. 
Our goal is to fully leverage and enhance the diverse contributions of these features towards localizing image inpainting.
To this end, we develop a new method for inpainting detection based on a Dense Feature Interaction Network (DeFI-Net). As shown in Table~\ref{tab:compare_sota}, while some methods have recognized the importance of multi-level features for inpainting detection, simplistic interaction mechanisms struggle to capture complementary correlations across levels. DeFI-Net uses a feature pyramid architecture to capture and amplify multi-scale representations across various stages, which can more effectively reveal the feature-level interactions for detecting image inpainting. This structure adaptively directs the lower-level features, which carry edge and shape information, to refine the localization of manipulated edges while integrating the higher-level semantic features. 
% Using DeFI-Net, we develop a method combining complementary representations to expose inpainted areas. 
Using DeFI-Net, we have developed a priori feature-guided edge attention module, incorporating inpainting edge prediction as an auxiliary task to enhance feature representation and improve localization accuracy at edge details. %that incorporates inpainting edge prediction as an  auxiliary task to enhance the model's ability to represent inpainting features and improve localization accuracy at edge details.} 
Evaluation on seven image inpainting datasets demonstrates the effectiveness of our approach, which achieves state-of-the-art performance in detecting inpainting across diverse models. The main contributions of our work can be summarized as follows:

\begin{itemize}
    % \item[$\bullet$] We introduce the DeFI-Net to highlight how layered features aid inpainting detection, with a dense feature pyramid architecture hierarchically and thoroughly encoding multi-view features through dense connections.
    \item[$\bullet$] We introduce the DeFI-Net to highlight how novel feature-level interactions aid inpainting detection, with a pyramid architecture hierarchically and thoroughly encoding multi-view features through dense connections.
    \item[$\bullet$] We further design a complementary supervision approach by adding edge supervision with reverse attention on early-stage features to the inpainting region localization using multi-level features.
    % \item[$\bullet$] To effectively integrate multi-level features, we design adaptive feature fusion modules to provide representative and compact context for identifying inpainting.
    \item[$\bullet$] To effectively integrate mid-low-level details and high-level semantics features, we design a spatial weight learning module that adaptively assigns fusion weights to provide representative and compact contexts for recognizing inpainting.
    \item[$\bullet$] Our method beats top algorithms in identifying traditional and DL-based inpainting, showing strong generalization ability to advanced inpainting techniques.
\end{itemize}

\begin{table*}[h!]
\renewcommand{\arraystretch}{1.0}
\caption{A comprehensive taxonomy of state-of-the-art methods for image inpainting detection.}
\scriptsize
\centering
\begin{tabular}{c|c|c|c|c|c|c|c}

\Xhline{0.8pt}
\multirow{2}{*}{\textbf{Methods}} & \multicolumn{2}{c|}{\textbf{Input information}} & \multirow{2}{*}{\textbf{Backbone}} & \multicolumn{3}{c|}{\textbf{Multi-level feature}} & \multirow{2}{*}{\textbf{Feature Interaction Mode}} \\ \cline{2-3} \cline{5-7} 
                      & RGB           & Noise        &                       & low-level & mid-level & high-level & \\ \hline
MT-Net~\cite{wu2019mantra}                & +  & \thead{BayerConv2D,\\SRM filters}   & VGG      & -  & -  & $\checkmark$ & - \\ \hline
HP-FCN~\cite{li2019localization}                & -             & High-pass filters    & FCN                   & -  & -  & $\checkmark$ & - \\ \hline
IID-Net~\cite{wu2021iid}               & +  & \thead{BayerConv2D,\\High-pass filters}     & Customized cell (NAS)                 & -  & -  & $\checkmark$ & - \\ \hline
PSCC-Net~\cite{liu2022pscc}              & +  & -            & HRNet                & $\checkmark$ & $\checkmark$ & $\checkmark$ & Progressive interaction \\ \hline%Top-down 
MVSS-Net~\cite{dong2022mvss}              & +  & BayerConv2D        & FCN                   & $\checkmark$ & $\checkmark$ & $\checkmark$ &  \thead{Dominated by \\high-level features} \\ \hline %Bottom-up
PS-Net~\cite{zhang2023ps}                & +  & \thead{SRM filters,\\First-order derivative filters, \\Second-order derivative filters} & Dense-Net           & -  & - & $\checkmark$ & - \\ \hline
DiffForensics~\cite{yu2024diffforensics}          & +  & -            & U-Net                 & $\checkmark$ & $\checkmark$ & $\checkmark$ & Skip connections \\ \hline %Top-down
CAT-Net~\cite{kwon2021cat}               & + & \thead{DCT coefficient,\\Quantization table} & HRNet  & $\checkmark$ & $\checkmark$ & $\checkmark$ &  \thead{HRNet-based fusion} \\ \hline %\thead{Cross-layer,\\Late-all-concat} 
TruFor~\cite{guillaro2023trufor}                & +  & Noiseprint++ & SegFormer             & $\checkmark$ & $\checkmark$ & $\checkmark$ & Late concatenation \\ \hline %Late-all-concat
DeFI-Net(Ours)                  & +  & -            & HRNet                & $\checkmark$ & $\checkmark$ & $\checkmark$ & \thead{Dense interaction\\ with context learning \\between layers} \\ %\thead{Top-down,\\Bottom-up,\\Cross-layer} \\ 
\Xhline{0.8pt}
\end{tabular}
\label{tab:compare_sota}
\end{table*}

\section{Related Work}
\subsection{Image inpainting manipulation}

Inpainting is a long-standing problem in image processing and computer vision~\cite{chu2023rethinking}. The task is ill-posed because there is no unique solution based on the remaining part of the image, and various prior information is needed to hallucinate the missing region. Early inpainting methods using non-learning algorithms aimed to recover lost pixels using pixel redundancy and local continuity priors, prioritizing speed and simplicity. Based on the principle of local continuity, Bertalmio et al.~\cite{bertalmio2001navier} proposed a method that extends isobaths using fluid dynamics, achieving seamless mixing in complex structures. The Fast Marching Method~\cite{telea2004image} employs a level-set approach to efficiently propagate pixel values inward, enabling smooth transitions in small, regular inpainting regions. Herling and Broll~\cite{herling2014high} developed a mapping function to establish correspondence between missing and original regions of an image, incorporating a composite cost function to ensure spatial and appearance consistency. To enhance structural coherence, Huang et al.~\cite{huang2014image} integrated mid-level cues such as planes and regularities, making their approach particularly effective for structured scenes with architectural elements. Additionally, Guo et al.~\cite{guo2017patch} applied Low-Rank Approximation to balance structural and textural recovery, utilizing low-rank decomposition to exploit image self-similarity and reconstruct both fine details and underlying structure. Collectively, these methods illustrate an evolution from basic pixel propagation techniques to advanced strategies incorporating structural integration, paving the way for more sophisticated inpainting tasks.

% Inpainting is a long-standing problem in image processing and computer vision~\cite{chu2023rethinking}. The task is ill-posed because there is no unique solution based on the remaining part of the image, and various prior information is needed to hallucinate the missing region. Early inpainting methods using non-learning algorithms aim to recover lost pixels using pixel redundancy and local continuity priors, either propagating information from the surrounding areas~\cite{bertalmio2000image} or using patch-based completion in a Markov random field model~\cite{ruvzic2014context}. 

More recent effective inpainting methods are obtained with deep neural networks (DNNs) trained on large datasets~\cite{xiang2023deep}, which allow for automatic feature learning and more context-aware inpainting results. Various types of DNN models have been explored, including the Generative Adversarial Network (GAN)~\cite{pathak2016context,iizuka2017globally}, Shift-Net~\cite{yan2018shift} used U-Net architecture, recent Fast Fourier Convolution models~\cite{chu2023rethinking, hu2023saliency, suvorov2022resolution, jain2023keys}, and decoupled probabilistic modeling~\cite{li2023image}. Learning-based methods emphasize adaptive feature selection and context integration to generate semantically coherent inpainting results. For example, Yu et al.~\cite{yu2019free} introduced gated convolutional layers to dynamically prioritize valid pixel information, enabling free-form inpainting across irregular mask shapes with high visual fidelity. Addressing challenges in large missing regions, where relying on neighboring pixel features alone is insufficient for accurate semantic reconstruction, Yu et al.~\cite{yu2018generative} introduced the attention mechanism that dynamically learns global contextual information, resulting in semantically consistent outputs. To further enhance texture consistency, a two-stage inpainting strategy was proposed by Wu et al.~\cite{learningdeep}, which combines a Local Binary Pattern network for structural estimation with spatial attention techniques. Additionally, to mitigate artificial artifacts and address mean and variance shift issues between missing and real areas, Yu et al.~\cite{yu2020region} introduced a region normalization process, which significantly improved inpainting quality.

The emergence of diffusion-based image generation models, such as DALL$\cdot$E 2~\cite{ramesh2022hierarchical} and Stable Diffusion~\cite{rombach2022high} further significantly improved the performance of image inpainting~\cite{yang2023uni, jia2023autosplice, corneanu2024latentpaint,xie2023smartbrush}. 
The diffusion-based image inpainting leads to more natural and realistic results that pose challenges to detection algorithms. 
%Investigating how to develop generalized solutions for detecting these low-cost and highly realistic image inpainting, deserves thorough exploration.

\subsection{Image inpainting detection}
Image inpainting detection is often solved with machine learning based on convolutional neural networks (CNNs)~\cite{mehrjardi2023survey}. These CNN-based approaches exploit hierarchical neural networks to automatically learn discriminative features for detecting inpainted regions.
Zhu et al. \cite{zhu2018deep} pioneered CNN-based image inpainting detection by employing a label matrix to guide automatic feature learning, and the weighted cross-entropy as the loss function. Several studies \cite{bappy2017exploiting, mazaheri2019skip, wu2019mantra} employed Long Short-Term Memory (LSTM) models to learn artifacts for manipulation localization.
HP-FCN~\cite{li2019localization} was proposed to enhance inpainting traces via high-pass filtering and learn discriminative features from image residuals. 
Similarly, IID-Net \cite{wu2021iid} used various combined filters to enhance inpainting features, and introduced a Neural Architecture Search (NAS) algorithm to adjust feature extraction cells.
These conventional encoder-decoder structures without pooling layers may hinder the model's capability to detect intricate tampering and result in spatial information loss. Several methods \cite{guillaro2023trufor, kwon2022learning, zhai2023towards, dong2022mvss} introduce multi-branch models to combine features from noise/frequency domains or patches~\cite{kong2023pixel}. New feature extraction models, including Transformer-based architectures \cite{li2023transformer, wang2022objectformer} or DenseNet-based approaches \cite{huang2017densely, zhang2023ps} have been used for inpainting detection. 
Furthermore, researchers have explored multi-scale feature learning techniques. For example, MVSS-Net~\cite{chen2021image, dong2022mvss} employed features learned from different stages of the backbone network and incorporated a two-branch supervision method considering RGB domain and noise patterns. Liu et al. \cite{liu2022pscc} introduced PSCC-Net, leveraging the multi-stage features learned from the HRNet~\cite{wang2020deep} for progressive mask prediction. 
GCA-Net \cite{das2022gca} designed a gated context attention network to capture fine image discrepancies in inpainted images. To improve global context representations and minimize feature loss, they refer to the UNet++~\cite{zhou2018unet++} architecture, which utilizes multi-scale bottom-up feature propagation in the decoding process. Recently, Yu et al. introduced DiffForensics~\cite{yu2024diffforensics}, which employs a self-supervised denoising diffusion paradigm integrated with an encoder-decoder architecture to detect and localize tampering. 

Unlike existing methods, this work highlights the complementary features of neural networks, and designs a dense feature interaction network tailored for capturing rich and discriminative patterns in image inpainting detection, as depicted in Fig. \ref{fig:module_compare}. 
{For the additional task of inpainting edge detection, edge extraction operator \cite{chen2021image, dong2022mvss, li2023edge} and ordinary convolutional structures \cite{zhou2020generate} are susceptible to interference from noise (shapes and boundaries of other authentic objects). Therefore, a priori feature-guided edge attention module is developed in this work to help the model focus on subtle differences inside and outside the boundary and improve the localization accuracy at  edges.}

\begin{figure*}%[htbp]
\centering
\includegraphics[width=0.99\linewidth]{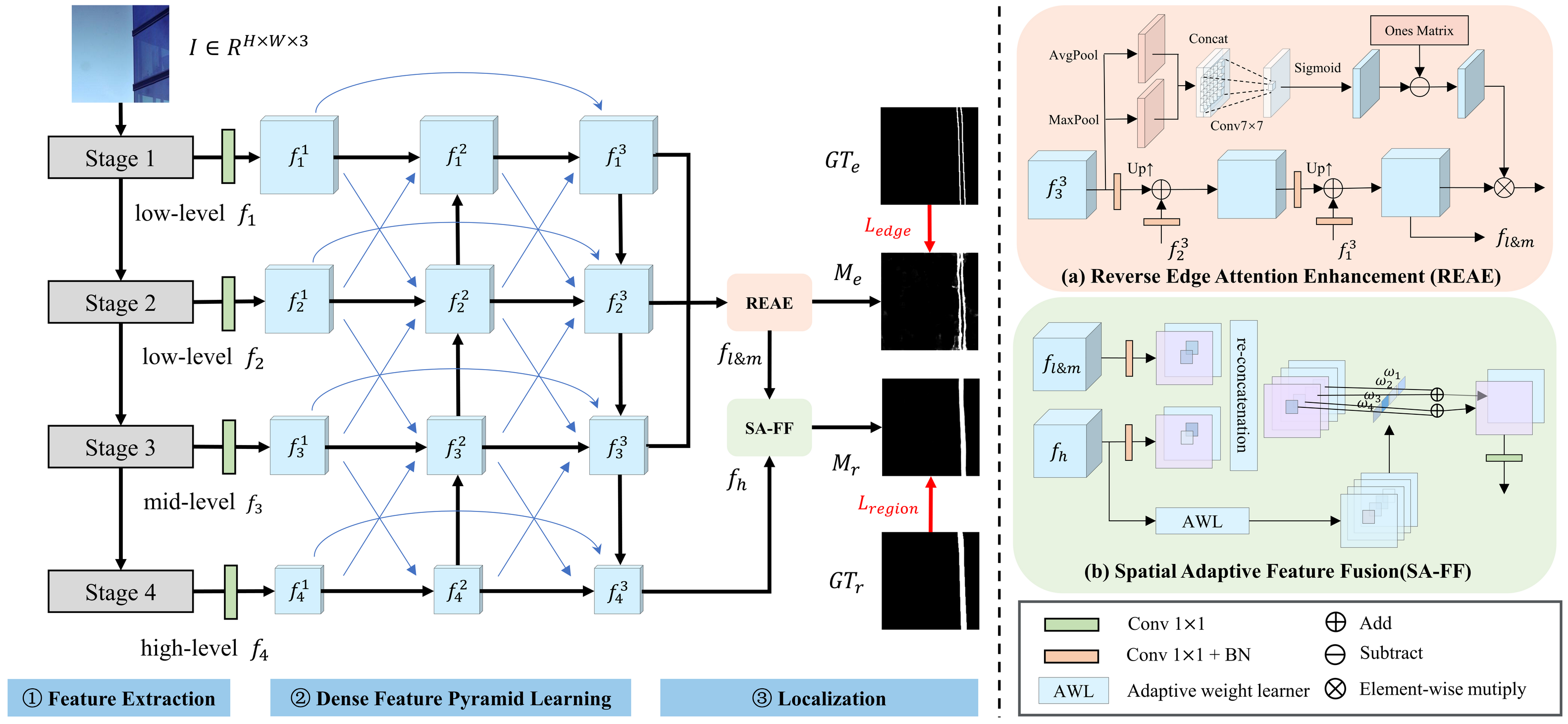}
\caption{Architecture of the proposed DeFI-Net. \textbf{Left}: The DFPL module densely integrates multi-stage features extracted from the backbone network. \textbf{Right}: Structure of REAE and SA-FF modules for inpainting artifact enhancement and multi-level semantic fusion.}
\label{fig:overall}
\end{figure*}

\section{DeFI-Net}
%\textbf{Architecture.} 
In this work, we introduce the DeFI-Net for image inpainting localization, as shown in Fig. \ref{fig:overall}. 
Taking an RGB image $I \in \mathbb{R}^{H\times W\times C}$ as input, the model will output a single-channel edge mask $M_e \in \mathbb{R}^{H\times W \times1}$ and region mask $M_r \in \mathbb{R}^{H\times W\times 1}$. The model begins with multiple stages of feature extraction from the backbone network. 
The multi-stage features are interacted and fused by three components: 1) a Dense Feature Pyramid Learning structure (DFPL) to introduce multi-scale features containing detailed texture, contextual and semantic information (\textit{see Sec.~\ref{sec:dfpl}}); 2) a Reverse Edge-Attention Enhancement (REAE), with mid-low-level features as a priori guiding the model to focus on edge inconsistencies caused by inpainting (\textit{see Sec.~\ref{sec:reae}}); {3) a Spatial Adaptive Feature Fusion (SA-FF) aims to learn the fusion weights of the high-level and mid-low-level features adaptively, unifying the representation of inpainting to achieve accurate detection results (\textit{ see Sec.~\ref{sec:saff}}).}

\subsection{Dense Feature Pyramid Learning}
\label{sec:dfpl}

Different layers in a Convolutional Neural Network (CNN) model have receptive fields of different sizes and extract image features of different levels of detail. 
Generally, early or shadow layers mostly extract low-level and generic image cues (e.g., edges and corners), and the middle layers carry shapes and textural patterns. In contrast, deeper layers capture high-level semantics or specific image cues~\cite{zeiler2014visualizing, mazaheri2019skip, lee2009convolutional}. 
Artifacts of image inpainting may be reflected at all three levels of features, each contributing in a complementary manner. For example, pixel filling can cause distortions or blur at edges, detectable by low-level features. Middle-level features mainly capture texture inconsistencies in the filled areas, like structural mismatches or color discordance. Alternatively, high-level features identify changes in an image's semantic content, such as missing object parts or altered scene interpretation, aiding the model in identifying inpainting areas. 

\begin{figure}[tbp]
\centering
\includegraphics[width=1\linewidth]{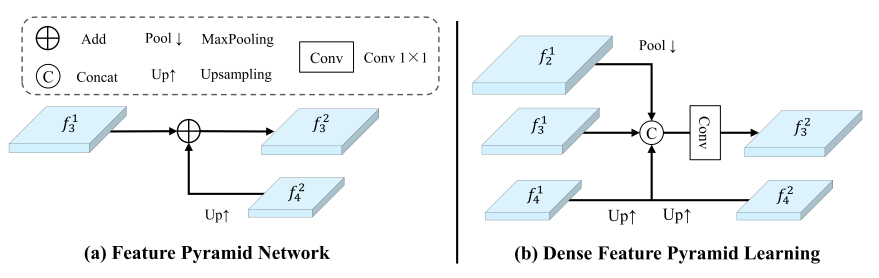}
\caption{Comparison of our DFPL with Feature Pyramid Network (FPN)~\cite{lin2017feature} in multi-level feature interaction. Note that the displayed interaction process corresponds to the top-down path, as described in Eq. (1) with $j=2$.}
\label{fig:dfpn}
\end{figure}

Our goal is to design an effective extraction and fusion of these complementary features (as shown in Fig.~\ref{fig:teaser}) to identify image inpainting. 
We first use the backbone to extract multi-stage features from the input image. Using a CNN (e.g., ~\cite{gao2019res2net}) divided into four stages as an example, we define the features from \textit{Stage 1} and \textit{Stage 2} as low-level features $f_1$ and $f_2$, the features extracted from \textit{Stage 3} as mid-level feature $f_3$, while the more abstract features from \textit{Stage 4} as high-level feature $f_4$. 
Inspired by the Feature Pyramid Network (FPN) \cite{lin2017feature} and Path Aggregation Network (PANet) \cite{liu2018path}, we design a multi-directional, multi-scale fusion method, as shown in Fig.~\ref{fig:dfpn}. In addition, we treat multi-level features as independent entities, emphasizing the specificity of features at different stages. Our densely connected fusion approach improves the efficiency of information interaction between adjacent and cross layers, progressively enhancing context awareness. %we treat multi-level features as independent entities, focusing more on the specificity of features at different stages. The densely connected fusion approach enhances the efficiency of information interaction between adjacent and cross layers, progressively enhancing context awareness.

Taking a single-scale feature as an example, our DFPL module considers the features of the same scale and all the neighboring features to fully exploit and enhance the useful context information for image inpainting localization. %In early-stage fusion, filtering valuable features is a more desirable approach. 
Specifically, to obtain the feature $f_i^j$ (where $i \in [1,4]$ denotes the stage of the backbone to which the current feature belongs, and $j \in [1,3]$ represents the state of the feature in DFPL, $j$ as 1 for the initial state involving channel reduction, 2 for the top-down path, and 3 for the bottom-up path), we concatenate its neighboring features in the channel dimension and use $1\times1$ convolution to integrate and select features. This process can be expressed by the equation below, where we have removed the max pooling, upsampling operations and cases with $i$ in the boundary condition for simplicity.

\begin{equation}
\label{eq:DFPN}
f_i^j = \left\{
                    \begin{array}{lr}
\text{Conv}_1(f_i), \quad\quad\quad\quad\quad\quad\quad\quad\quad\quad\quad \text{if}  \,\,j = 1,\\           
\text{Conv}_1(\text{Cat}(f_{i-1}^{1}, f_{i}^{1}, f_{i+1}^{1}, f_{i+1}^{2})),  \:\:\:\:\:\:\:\: \text{if} \,\, j = 2, \\%i\in [2,4] &  \\
\text{Conv}_1(\text{Cat}(f_i^{1}, f_{i-1}^{2}, f_{i}^{2}, f_{i+1}^{2}, f_{i-1}^{3})),  \:\: \text{if}\,\, j = 3.%, i\in [2,4].  &  \\
                    \end{array}
        \right.
\end{equation}
$Cat(\cdot)$ refers to concatenating the required features in the channel dimension, and $Conv_1(\cdot)$ represents $1\times1$ convolution layer. %According to our feature grading strategy, $i = 1, 2, 3, 4,$ or $5$. 
% \noindent where $j = 2$ represents the bottom-up path, while $j = 3$ represents the top-down path.

\subsection{Reverse Edge Attention Enhancement}
\label{sec:reae}
Filling content to a specific area in an image can leave detectable traces between the altered region and its surroundings. Therefore, leveraging these edge artifacts is crucial for detecting inpainting, which the {mid-low-level} features can guide. {Nevertheless, mid-low-level features not only contain inconsistencies caused by inpainting but also a large amount of detail and structural noise, such as texture, shape, and object boundaries. Using traditional edge extraction operators \cite{dong2022mvss, li2023edge} or convolutional structures \cite{zhou2020generate} will weaken the representation of inpainting artifacts. Leveraging context-rich features that include the structural information of the restoration region as a priori can effectively suppress irrelevant semantic noise and guide boundary learning. %Utilizing features enriched with context and encompassing the structural information of the restoration region as a prior can effectively suppress irrelevant semantic noise, thereby guiding boundary learning.} %Boundary learning guided by rich context priors can effectively overcome these negative impacts.

We develop a method for detecting and enhancing edge artifacts using reverse attention called REAE, which uses the structural details in the mid-level feature $f_3^3$ as a guide, directing the attention of low-level features toward subtle boundary changes of the inpainting area. This promotes learning distinctive features to differentiate between inpainting edges and natural boundaries. As shown in Fig.~\ref{fig:overall}(a), the REAE module accepts mid-level feature $f_3^3$ and low-level features $f_1^3$ and $f_2^3$ as input. Its upper branch aggregates structural information of suspicious regions through average pooling, max pooling, and $7 \times 7$ convolution. Subsequently, the sigmoid output is inverted to obtain a reverse attention map emphasizing edges, expressed by Eq. (\ref{eq:REAM_1}). Meanwhile, the lower branch of the REAE progressively combines $f_3^3$, $f_2^3$, and $f_1^3$ for $f_{l\&m}$ through $3 \times 3$ convolution, Batch Normalization (BN), and bilinear interpolation upsampling. Finally, we use Eq. (\ref{eq:REAM_2}) to fuse the feature for edge supervision {and output the edge mask $M_e$ through a prediction head composed of multiple convolutional layers}.

%\vspace{-0.6cm}
\begin{align}
\label{eq:REAM_1}
Att_{re} &= 1 - Sigmoid(Conv_7(Cat(Pool_{avg}(f_3^3), \notag \\
&\phantom{{}=}Pool_{max}(f_3^3))))
\end{align}

\begin{equation}
\label{eq:REAM_2}
%\vspace{-0.1cm}
Me = Pred_{edge}(f_{l\&m} \otimes Att_{re}),
\end{equation}

\noindent where $Conv_7(\cdot)$ denotes a $7\times7$ convolution layer. $Pool_{avg}$ and $Pool_{max}$ are defined as average pooling and max pooling, respectively. $Pred_{edge}$ is the prediction head, composed of two convolutional layers with $3\times3$ and $1\times1$ kernels, which outputs a single-channel edge mask.

\begin{figure}[tbp]
\centering
\includegraphics[width=9cm]{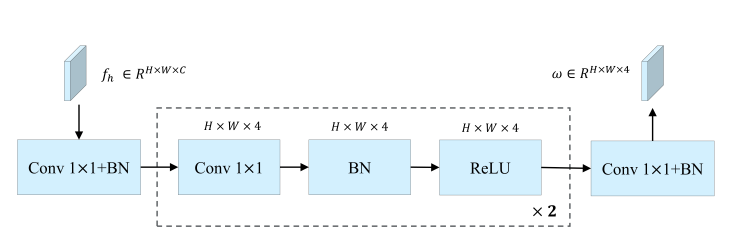}
\caption{{Structure of the adaptive weights learner. Assigning reasonable fusion weights to different spatial locations by successive $1\times1$ convolutions.}}
\label{fig:awl}
\end{figure}

\subsection{Spatial Adaptive Feature Fusion}
\label{sec:saff}
{Due to the nature of neural networks, high-level feature maps exhibit a significantly stronger response to the target region compared to low-level features. The difference in receptive field inevitably leads the model to bias towards the side with a more robust response during the fusion process, allocating more weights to the high-level features, which can severely suppress the role of low-level features. The latter can provide detailed edge positioning and structural information, significantly impacting the detection accuracy.}

{To unify the feature representation of the inpainting region, we design a spatially adaptive feature fusion module (SA-FF) to combine high-level features and mid-low-level features, correcting the bias brought by response intensity. Normalization can limit the response of different features to a similar magnitude, effectively eliminating the interference of response intensity. After normalization, the model can treat multi-level features fairly, making it easier to learn effective fusion weights. Specifically, we reduce the channels of $f_{l\&m}$ and $f_{h}$ to 2 through a $1\times1$ convolution, followed by normalization. The process can be expressed as: Eq. (\ref{eq:Norm_1}) and (\ref{eq:Norm_2}).

\begin{equation}
\label{eq:Norm_1}
%\vspace{-0.1cm}
F_{l\&m}^{Norm} = BN(Conv_{1}(f_{l\&m})), F_{l\&m}^{Norm} \in \mathbb{R}^{H\times W\times 2}
\end{equation}
\begin{equation}
\label{eq:Norm_2}
%\vspace{-0.1cm}
F_{h}^{Norm} = BN(Conv_{1}(f_{h})), F_{h}^{Norm} \in \mathbb{R}^{H\times W\times 2}
\end{equation}

{For different spatial positions, there exists a disparity in the contribution of high-level and mid-low-level features. High-level features, rich in context, are semantically accurate but lose many spatial and geometric details. Mid-low-level features better retain these aspects, and we hope that the model will rely more on the latter in boundary areas while leveraging the information provided by the former to fill the target area. As shown in the Fig.~\ref{fig:awl}, we designed a spatially adaptive weight learner to adjust the fusion weights according to positional features, fully utilizing both advantages. Specifically, we feed $f_{h}$ used for learning weights into alternating connection blocks composed of $1\times1$ convolution, BN, and ReLU to generate position-aware fusion weights $\textbf{W}\in\mathbb{R}^{H\times W\times 4}$. Then, the final fusion output is obtained according to element-wise multiplication and channel-wise addition, as shown in Eq. (\ref{eq:fuse_2}-\ref{eq:fuse_4}). Finally, a region mask $M_r$ is output through a prediction head composed of convolutional layers.}

\begin{equation}
\label{eq:fuse_1}
\begin{aligned}
\vspace{-0.3cm}
\textbf{W} &= ((\omega_1)_{i,j}, (\omega_2)_{i,j}, (\omega_3)_{i,j}, (\omega_4)_{i,j}), 
\\ &i\in[1,H] ,j\in[1,W]
\end{aligned}
\end{equation}

\begin{equation}
\label{eq:fuse_2}
\vspace{-0.3cm}
F^{Norm} = Cat(F_{h}[0], F_{l\&m}[0], F_{h}[1], F_{l\&m}[1])
\end{equation}

\begin{equation}
\label{eq:fuse_3}
\begin{aligned}
\vspace{-0.3cm}
\mathbf{F}^{\text{Fuse}}[k] & = (\mathbf{W}[2k] \otimes \mathbf{F}^{\text{Norm}}[2k]) \oplus (\mathbf{W}[2k+1] \\ & \otimes \mathbf{F}^{\text{Norm}}[2k+1]), k \in [0,1]
\end{aligned}
\end{equation}

\begin{equation}
\label{eq:fuse_4}
\vspace{-0.1cm}
\mathbf{F}^{\text{Fuse}} = \text{Cat}\left(\mathbf{F}^{\text{Fuse}}[0], \mathbf{F}^{\text{Fuse}}[1]\right)
\end{equation}

\noindent {where $Cat(\cdot)$ represents the ordinary concatenation along the channel dimension. $\otimes$ and $\oplus$ denote the element-wise multiplication and element-wise addition, respectively. $[\cdot]$ is defined as the slicing operation of the channel dimension. Note that in our design, \textbf{W} will learn different weights for each spatial position of each channel, totaling $4\times H\times W$ weights.}

\subsection{Loss Function}
To train our DeFI-Net, we use an objective function that combines the region loss and the edge loss. The region loss enhances the model’s sensitivity to the inpainting region globally, promoting the overlap between the predicted mask and the ground truth. In contrast, the edge loss guides the model to learn non-semantic detail differences from a local perspective, achieving more precise localization. {According to \cite{wei2020f3net}}, we define the region loss function as $L_{region} = L_{bce}^w + L_{iou}^w$. Compared with the traditional Binary Cross-Entropy (BCE) loss, $L_{bce}^w$ assigns different weights to pixels based on their positions, paying more attention to the target area and hard pixels (e.g., boundaries prone to classification errors). The IoU loss $L_{iou}^w$ assigns greater weight to the target area, encouraging the network to focus on the global structure. For the edge loss, we use $L_{bce}^w$, similar to supervise the edge prediction, denoted as $L_{edge}$. The overall loss function is a convex combination of two losses: $ L_{total}=\lambda_r L_{region} + \lambda_e L_{edge} $. 
% Further details on the loss functions are available in the Supplementary Material.

\section{Experiments}
\label{sec:exper}

\subsection{Experimental Settings}

\noindent\textbf{Datasets.} 
To evaluate the performance of our DeFI-Net, we use seven diverse image inpainting datasets, considering variations in inpainting model, region size, and image content. Table \ref{tab:data} summarizes the datasets. IID-84K \cite{wu2021iid} comprises 74K inpainted images created using the Gated Convolution (GC) model \cite{yu2019free} on images sourced from Places \cite{zhou2017places} (with JPEG lossy compression) and Dresden \cite{gloe2010dresden} (with NEF lossless compression). Additionally, it includes a subset with 1K GC images for in-domain testing (denoted as IID-GC) and another 9K images (denoted as IID-9K) generated by four other deep learning inpainting models (CA~\cite{yu2018generative}, SH~\cite{yan2018shift}, LB~\cite{learningdeep}, RN~\cite{yu2020region}) and five traditional inpainting models (TE~\cite{telea2004image}, NS~\cite{bertalmio2001navier}, PM~\cite{herling2014high}, SG~\cite{huang2014image}, LR~\cite{guo2017patch}) on images from CelebA \cite{karras2017progressive} and ImageNet \cite{deng2009imagenet}. DEFACTO~\cite{mahfoudi2019defacto} contains different manipulation types while our primary focus lies on 20K inpainting images and 4K authentic images (DE-24K) for training, along with a validation set (DE-1K) and a test set (DE-3K). In addition, we consider diverse datasets created by modern photo-editing software (e.g., Korus~\cite{Korus2016TIFS}, Photoshop~\cite{zhang2023ps}) and AIGC technology (e.g., AutoSplice~\cite{jia2023autosplice}, CocoGlide~\cite{guillaro2023trufor}, TGIF~\cite{mareen2024tgif}) to evaluate the generalization ability of our method.

\setlength{\tabcolsep}{1.3pt} 
\begin{table}[h]
\renewcommand{\arraystretch}{1.1}
\centering
\caption{Details of image inpainting datasets in our experiment.}
\newcommand{\tabincell}[2]{\begin{tabular}{@{}#1@{}}#2\end{tabular}} 
\scalebox{0.76}{
\begin{tabular}{c|c|c|c|c|c}
\Xhline{0.8pt}
\multirow{2}{*}{\textbf{Dataset}} & \multicolumn{2}{c|}{\textbf{Training}}       & \multicolumn{2}{c|}{\textbf{Testing}}        & \multirow{2}{*}{\textbf{Data Source}} \\ \cline{2-5}
                         & \multicolumn{1}{c|}{\textbf{\#Sample}} & \textbf{Model} & \multicolumn{1}{c|}{\textbf{\#Sample}} & \textbf{Model} &                              \\ \hline    
IID-84K~\cite{wu2021iid}    & {74K}      & {GC}     & {10K}                & \tabincell{c}{GC, CA, SH,\\LB, RN, TE,\\NS, PM, SG, LR}   &\tabincell{c}{Places~\cite{zhou2017places}, Dresden~\cite{gloe2010dresden}, \\CelebA~\cite{karras2017progressive}, ImageNet~\cite{deng2009imagenet}}   \\ \hline
DEFACTO\cite{mahfoudi2019defacto}     & {24K}        & Exemplar {\cite{daisy2014smarter}}         & {3K}                 &   Exemplar {\cite{daisy2014smarter}}      & MSCOCO~\cite{lin2014microsoft}         \\ \hline
{Korus}~\cite{Korus2016TIFS}      & {-}        & -           & {57}      &    \tabincell{c}{GIMP, \\Affinity softwares} &    \tabincell{c}{Sony, Canon, \\Nikon Photos}      \\ \hline
{Photoshop}~\cite{zhang2023ps}  & {-}       & -             & {300}     & Photoshop     & Places~\cite{zhou2017places}                       \\ \hline
{AutoSplice}~\cite{jia2023autosplice}  & {-}      & -               & {3.6K}       & DALLE-2    &   Visual News \cite{liu2020visual}    \\ \hline
CocoGlide~\cite{guillaro2023trufor}  & {-}      & -               & 512     & GLIDE    &   MSCOCO~\cite{lin2014microsoft}    \\ \hline
TGIF~\cite{mareen2024tgif}  & {-}      & -               & 4.1K       & \thead{SD2, SDXL, \\Adobe Firefly}   &   MSCOCO~\cite{lin2014microsoft}    \\ 
\Xhline{0.8pt}
\end{tabular}}
\label{tab:data}
\end{table}

\noindent\textbf{Evaluation Metrics.} Our evaluation metrics include the Area Under the Receiver Operating Characteristic Curve (AUC), F1 score, and Intersection over Union (IoU). AUC and F1 measure the accuracy of binary classification (authentic or inpainted) at the pixel level, while IoU quantifies the overlap between the predicted region and the ground truth. These three metrics have a range of [0, 100] percent, where higher values indicate better performance. We use a threshold of 0.5 for F1 and IoU evaluation.

\noindent\textbf{Implementation Details.} 
We use the HRNet~\cite{wang2020deep} initialized with pre-trained weights from ImageNet~\cite{deng2009imagenet} as an example. Our model is implemented in Pytorch and trained using the Adam optimizer. During the training phase, we set the size of the input image to $352\times352$ with standard data augmentation
techniques, including random flipping, cropping, and rotation. The batch size is 16. The initial learning rate is 2e-4 and will be halved if the AUC on the validation set fails to increase for five epochs until the convergence. In the DFPL module, the channels of all-level features are reduced to $64$.  In addition, we set the $\lambda_r$ and $\lambda_e$ for the region and edge loss to 5 and 1, respectively. 

\subsection{{Quantitative Comparisons}}

\noindent \textbf{In-domain Localization}. We first compare our DeFI-Net with several open-source methods, including MT-Net~\cite{wu2019mantra}, HP-FCN~\cite{li2019localization}, IID-Net~\cite{wu2021iid}, PS-Net~\cite{zhang2023ps}, PSCC-Net~\cite{liu2022pscc}, MVSS-Net~\cite{dong2022mvss}. For fairness, we train our method and all baselines using the same settings on the IID-74K and DE-24K datasets, respectively. Table~\ref{tab:1} presents the localization results under in-domain testing. Our method achieves the best performance on both datasets. Specifically, our AUC, F1, and IoU scores are $1.07\%$, $2.83\%$, and $3.76\%$ higher than those of the second-place method, respectively. On the DE-3K dataset, single-scale encoding-decoding methods (such as IID-Net) experienced a notable performance drop due to the challenges in identifying small inpainting regions in this dataset. Our DeFI-Net outperforms the comparison methods significantly, particularly achieving a 23.3\% lead over the second-place PSCC-Net in terms of the IoU metric. 
This demonstrates the effectiveness and superiority of our dense feature pyramid learning design in facilitating sufficient interaction among multi-scale features, thereby enhancing the detection ability even for small inpainting targets.

% \usepackage{multirow}
\definecolor{lightgray}{rgb}{0.9, 0.9, 0.9}
\setlength{\tabcolsep}{1.6pt} 
\begin{table}[t]
\renewcommand{\arraystretch}{1.1}
\centering
\caption{Quantitative comparison results (\%) under in-domain testing on IID-GC and DE-3K datasets.}
\newcommand{\tabincell}[2]{\begin{tabular}{@{}#1@{}}#2\end{tabular}} 
\label{tab:1}
\scalebox{0.9}{
%\begin{tabular}{cccccccc{\columncolor{lightgray}}}
\begin{tabular}{c|c|cccccc>{\columncolor{lightgray}}c}
\Xhline{0.8pt}
\multirow{2}{*}{Dataset} & \multirow{2}{*}{Metric} &  \tabincell{c}{MT-Net\\ \cite{wu2019mantra}} & \tabincell{c}{HP-FCN\\ \cite{li2019localization}} &\tabincell{c}{IID-Net\\ \cite{wu2021iid}} & \tabincell{c}{PSCC-Net\\ \cite{liu2022pscc}} & \tabincell{c}{MVSS-Net\\ \cite{dong2022mvss}} & \tabincell{c}{PS-Net\\ \cite{zhang2023ps}} & \tabincell{c}{DeFI-Net\\(Ours)} \\ \hline
IID-GC              & AUC                     & 93.83                      & 97.05                      & 97.54                      & 97.53                        & 85.09                        & 91.19                      & \textbf{98.61}           \\
IID-GC                & F1                & 63.62                      & 80.07                      & 89.08                       & 88.36                        & 63.55                        & 88.07                      & \textbf{91.91}           \\
IID-GC               & IoU                     & 50.73                      & 70.19                      & 77.89                       & 79.90                        & 26.35                        & 75.52                      & \textbf{83.66}           \\ \hline
DE-3K           & AUC                     & -                          & -                          & 55.98                       & 72.84              & 62.98           & 68.27                      & \textbf{79.36}           \\
DE-3K           & F1               & -                          & -                          & 49.85                       & 58.01                        & 60.09                        & 45.47                      & \textbf{71.62}           \\
DE-3K           & IoU                     & -                          & -                          & 1.50                        & 13.78                        & 4.87                         & 8.10                       & \textbf{37.08}           \\ \Xhline{0.8pt}
\end{tabular}}
\end{table}

\noindent \textbf{Generalized Localization}. To assess the generalization capability of the proposed DeFI-Net in localizing unknown inpainting techniques and types, we conducted a comparison. This involved applying the methods trained on the GC model to detect nine unseen inpainting models in the IID-9K dataset. The results in Table~\ref{tab:2} demonstrate that our method achieves the highest performance across the averaged AUC, F1 score, and IoU metrics. It is worth noting that the PSCC-Net exhibits slightly better performance than our methods in detecting image inpainting from the {LB, PM, SG, and LR models}, {indicating its generalizability benefited from the coarse to fine-grained multilevel supervision and progressive feature fusion strategy. We also note the latest diffusion-based detection algorithm~\cite{yu2024diffforensics} and further compare our method with baselines on more diverse image inpainting models.}
% indicating its generalizability benefited from the HRNet \cite{wang2020deep} feature extractor and progressive feature fusion strategy. We further compare our method with baselines on more diverse image inpainting models and also evaluate its performance against the HRNet extractor. 
Table~\ref{tab:3} shows that our dense feature network exhibits strong generalization capabilities in localizing various inpainting models, including software-based and generative AI models. 
% Our approach with HRNet further improve the performance in detecting inpainted images in all testing datasets. %This discrepancy may stem from the large inpainting regions in this dataset (even with entire image synthesis), for which our multi-scale context analysis may be limited.

Notably, DeFI-Net shows a significant performance gap compared to CAT-Net and TruFor on the TGIF dataset. During the construction of TGIF, the generated inpainted regions were spliced back into the original image using the provided masks and an additional small gradient boundary (a few pixels wide), effectively reducing artifacts. CAT-Net and TruFor incorporate additional noise-based features to aid detection, enabling them to capture subtle noise discrepancies between the generated content and the original image. In contrast, DeFI-Net primarily relies on detail differences (e.g., texture, color, edges) and global inconsistencies to identify inpainted regions, leading to reduced performance when dealing with such specific processing. In future work, we plan to explore robust noise representations to further enhance the model's generalization capability.

\noindent \textbf{Robustness Evaluation}. 
Inpainted images will inevitably undergo various post-processing operations when disseminated. Malicious editors may also use post-processing to conceal operation traces. Thus, robustness is of vital significance. To evaluate the model's robustness to post-processing, we consider five common operations: JPEG compression, adding Gaussian noise, adding salt and pepper noise perturbation, blurring, and image resizing. For each post-processing method, we conducted a comprehensive evaluation of the quality of JPEG compression (from 85 to 100), the standard deviation of Gaussian noise (from 0 to 1.5), the proportion of Salt\&Pepper noise (from 0 to 0.15), the kernel size of Gaussian blur (from 1 to 7), and the scaling factor for image resizing (ranging from 100\% to 85\% of the original size). The results obtained from employing all pre-trained models on post-processed IID-9K data are presented in Fig~\ref{fig:roub}. We observe that our DeFI-Net exhibits great robustness to noise perturbation, JPEG compression, and slight image resizing. Yet, MVSS-Net, PS-Net, and IID-Net rely on features from the noise stream, resulting in a weak anti-interference capability.} The performance of all methods deteriorates on Gaussian blurred images. This is because blurring eliminates typical features left by the inpainting in the boundary region, effectively concealing the inpainting operation. However, the AUC of DeFI-Net still outperforms the baseline by 5\% to 8\%.

\begin{table*}[t]
\centering
\caption{Quantitative comparison results (\%) under cross-domain testing on IID-9K dataset. The best performance in each column is highlighted in \textbf{bold}, while the second highest value is \underline{underlined}.}
\label{tab:2}
\scalebox{1}{
\begin{tabular}{r|c|cccc|ccccc|c}
\Xhline{0.8pt}
\multirow{2}{*}{Model} & \multirow{2}{*}{Metric}   & \multicolumn{4}{c|}{DL-based Inpainting}                                   & \multicolumn{5}{c|}{Traditional Inpainting}                                                 & \multirow{2}{*}{Average} \\ \cline{3-11}
             &             & CA~\cite{yu2018generative}             & SH~\cite{yan2018shift}     & LB~\cite{learningdeep}             & RN~\cite{yu2020region}& TE~\cite{telea2004image}  & NS~\cite{bertalmio2001navier}    & PM~\cite{herling2014high}           & SG~\cite{huang2014image}            & LR~\cite{guo2017patch}&                       \\ \hline
MT-Net \cite{wu2019mantra}                 & \multirow{7}{*}{AUC}      & 87.46          & 97.67          & 97.54          & 93.75                   & 91.36          & 80.12          & 95.36          & 98.42          & 89.15                   & 92.31                 \\
HP-FCN \cite{li2019localization}                &                           & 92.41          & 93.54          & 84.98          & 97.44                   & 89.74          & \underline{ 95.91}    & 98.23          & 99.85          & \textbf{98.90}          & 94.56                 \\
IID-Net \cite{wu2021iid}                &                           & 95.26          & 99.02          & \underline{99.23}          & 97.95                   & \underline{94.85}     & 94.79          & \underline{99.15} & \textbf{99.90} & \underline{98.75}              & 97.66                 \\
PSCC-Net \cite{liu2022pscc}               &                           & \underline{ 96.63}    & \underline{99.62} & \textbf{ 99.57}    & \underline{ 99.04}             & 94.67          & 95.27          & 99.12          & \underline{99.88}     & 98.35         & \underline{98.02}           \\
MVSS-Net \cite{dong2022mvss}                &                           & 93.06          & 94.68          & 95.03          & 90.90                   & 94.77          & 94.57          & 82.39          & 97.17          & 83.85                   & 91.82                 \\
PS-Net \cite{zhang2023ps}                &                           & 83.99          & 92.55          & 91.80          & 91.32                   & 85.51          & 82.83          & 76.62          & 91.50          & 85.36                   & 86.83                 \\
      \rowcolor{lightgray}
        DeFI-Net (Ours)                &                           & \textbf{97.26} & \textbf{99.68}    & 99.19  & \textbf{99.29}          & \textbf{98.10} & \textbf{96.56} & \textbf{99.17}     & 99.86          & 98.65                  & \textbf{98.64}        \\ \hline
MT-Net \cite{wu2019mantra}               & \multirow{7}{*}{F1} & 61.05          & 77.45          & 83.88          & 65.00                   & 61.46          & 35.92          & 37.71          & 70.32          & 29.32                   & 58.01                 \\
HP-FCN \cite{li2019localization}                &                           & 69.36          & 66.05          & 42.09          & 79.50                   & 64.81          & 78.41          & 52.95          & 88.51          & 69.16                   & 67.87                 \\
IID-Net \cite{wu2021iid}               &                           &  \underline{84.98}    & 92.90          &   \underline{93.22}         &  82.87                   &  \underline{85.03}    &  81.80          & 77.28          &  \textbf{95.86}          & 78.75                   & 85.85                 \\
PSCC-Net \cite{liu2022pscc}               &                           & 83.41          & \underline{93.91}  &  \textbf{94.54}    &  \underline{92.66}             &  82.34          &  83.39    &  \textbf{85.08}  & 95.04  & \textbf{88.01}          & \underline{88.71}           \\
MVSS-Net \cite{dong2022mvss}               &                           & 73.28          & 
 80.15          &  72.09          & 68.36                   &  84.17          & \underline{84.18}          &  54.46         &  75.77          &  60.33                   &  72.53                 \\
PS-Net \cite{zhang2023ps}                &                           & 83.24          & 93.44          &  93.19          &  92.36                   &  84.06          & 82.16          & 80.54          &  93.14          &  \underline{87.57}                   &  87.74                 \\
      \rowcolor{lightgray}
DeFI-Net (Ours)                &                           & \textbf{89.65} & \textbf{ 95.54}    &  92.42   & \textbf{94.08}          & \textbf{89.81} & \textbf{86.69} & \underline{81.99}    & \underline{95.22}    & 83.38             & \textbf{89.87}        \\ \hline
MT-Net \cite{wu2019mantra}               & \multirow{7}{*}{IoU}      & 42.29          & 65.76          & 74.57          & 49.98                   & 48.39          & 24.22          & 24.57          & 57.90          & 18.51                   & 45.13                 \\
HP-FCN \cite{li2019localization}                &                           & 56.82          & 51.34          & 29.08          & 67.38                   & 51.31          & 67.73          & 37.99          & 79.99          & 55.49                   & 55.24                 \\
IID-Net \cite{wu2021iid}           &                           & 67.77          & 80.90          & \underline{82.30}          & 58.80                   & 69.36          & 61.11          & 41.21          & \textbf{85.86}     & 47.56                   & 66.10                 \\
PSCC-Net \cite{liu2022pscc}               &                           & \underline{72.60}  &  \underline{83.01}  & \textbf{84.87}    &  \underline{80.22}             & \underline{69.90}    &  \underline{71.82}    &  \textbf{56.52}  & 83.15  & \textbf{65.58}          & \underline{74.19}           \\
MVSS-Net \cite{dong2022mvss}                &                           & 42.94          & 55.59          & 41.98          & 34.32                   & 63.22          & 63.05          & 6.43           & 43.76          & 16.32                   & 40.85                 \\
PS-Net \cite{zhang2023ps}                &                           & 63.83          & 81.49          & 81.51          & 79.01                   & 66.94          & 61.27          & 46.13          & 77.57          & \underline{ 64.66}             & 69.16                 \\
      \rowcolor{lightgray}
DeFI-Net (Ours)                &                           & \textbf{77.73}    & \textbf{87.10}    &  80.13  & \textbf{83.59}          & \textbf{79.11} & \textbf{71.87} & \underline{ 50.29}    & \underline{84.01}          & 56.82                   & \textbf{74.52}        \\ \Xhline{0.8pt}
\end{tabular}}
\end{table*}

\setlength{\tabcolsep}{1.8pt} 
\begin{table*}[t]
\centering
\caption{Quantitative comparison ({F1} \%) under cross-domain testing on diverse datasets. The best performance in each column is highlighted in \textbf{bold}, while the second highest value is \underline{underlined}. \dag \enspace denotes the result of using open source pre-trained weights. * indicates that the result is referenced from the paper \cite{yu2024diffforensics}.}
\newcommand{\tabincell}[2]{\begin{tabular}{@{}#1@{}}#2\end{tabular}} 
\label{tab:3}
\scalebox{1.1}{
\begin{tabular}{r|c|cccccc|c}
\Xhline{0.8pt}
Model   & \tabincell{c}{Training Data\&Size}    & \tabincell{c}{AutoSplice  \cite{jia2023autosplice}}    & \tabincell{c}{Korus \cite{Korus2016TIFS}}          & \tabincell{c}{Photoshop  \cite{zhang2023ps}}     & \tabincell{c}{DE-3k  \cite{mahfoudi2019defacto}}   & \tabincell{c}{CocoGlide}  \cite{mahfoudi2019defacto}     & \tabincell{c}{TGIF}  \cite{mahfoudi2019defacto}    & Average           \\ \hline
IID-Net \cite{wu2021iid}  & IID-74K    & 55.40          & 33.98          & 46.66          & 27.17          & 56.74  & 40.16  & 43.35          \\
PSCC-Net \cite{liu2022pscc} & IID-74K      & 61.49   & 26.17           & 58.77    & 31.33   & 52.63    & 36.58  & 44.49  \\
MVSS-Net \cite{dong2022mvss}  & IID-74K          & 41.07          & 42.06    & 55.18          & 41.68          & 50.55   & 39.56     & 45.01         \\
PS-Net \cite{zhang2023ps}  & IID-74K              & 46.13          & 44.15          & 49.20         & 33.76          & 49.18      & 38.16       & 43.43          \\ \hline
CAT-Net\textsuperscript{\dag}\cite{kwon2022learning}  & CASIAv2,etc -870K     & \underline{68.50}          & \underline{48.46}       & \underline{64.12}       & \textbf{50.69}       & 55.20     & \textbf{64.00}     & \textbf{58.49}         \\
TruFor\textsuperscript{\dag}\cite{guillaro2023trufor}      & CASIAv2,etc -870K   &\textbf{72.63}       & 48.05       & 47.83       & 42.34       & \textbf{62.64}     & \underline{50.12}     & 53.93       \\
DiffForensics*\cite{yu2024diffforensics}    & CASIAv2,etc -49K          & 50.70          & 25.70          & -          & -          & -     & -     & -          \\ \hline
%      \rowcolor{lightgray}
%DeFI-Net (Ours-Res2Net)           & 76.22        & \underline{57.00}        & \underline{94.78} &  60.28    & 72.07    \\ 
      \rowcolor{lightgray}
DeFI-Net (Ours)   & IID-74K        & 67.96    & \textbf{48.89} & \textbf{77.94} & \underline{43.88}    & \underline{57.13}   & 40.95     & \underline{56.13}    \\ \Xhline{0.8pt}
\end{tabular}}
\end{table*}

\def\subFigSz{0.195\linewidth}
\def\subFigH{0.15\linewidth}
\def\subCaptionSzA{0.23\linewidth}
\def\subCaptionSzB{0.16\linewidth}
\def\subCaptionSzC{0.22\linewidth}
\def\subCaptionSzD{0.19\linewidth}
\def\subCaptionSzE{0.18\linewidth}
% \captionsetup{type=figure} 
% \includegraphics[]{
\begin{figure*}[t]
\centering
\begin{minipage}[c]{0.99\linewidth}
    \includegraphics[width=\subFigSz, height=\subFigH]{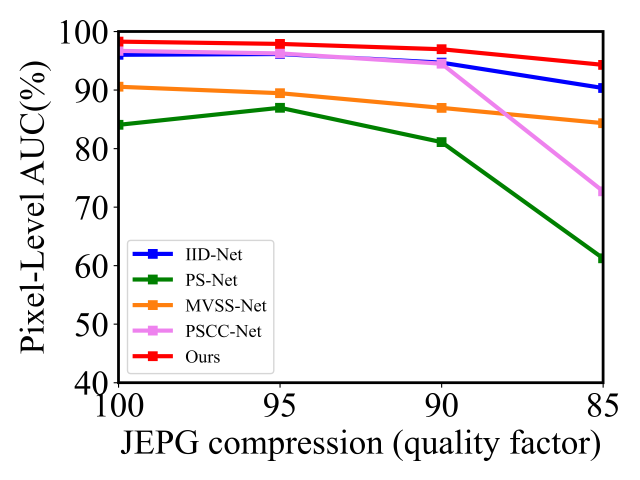}\hspace{0pt}
    \includegraphics[width=\subFigSz, height=\subFigH]{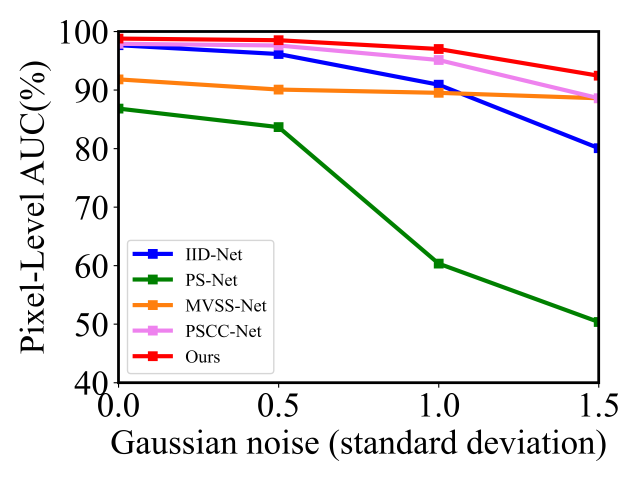}\hspace{0pt}
    \includegraphics[width=\subFigSz, height=\subFigH]{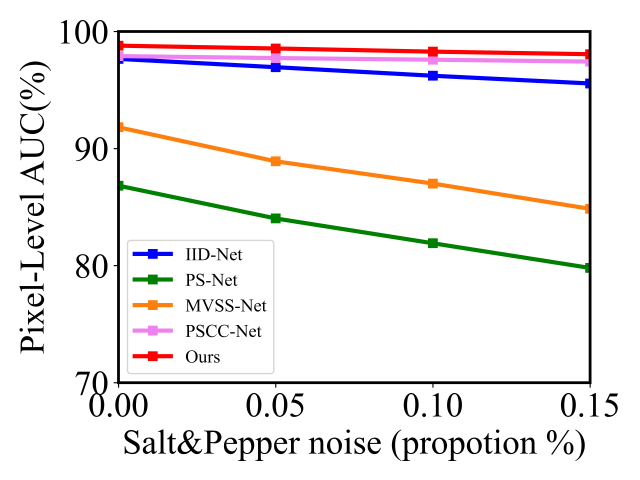}\hspace{0pt}
    \includegraphics[width=\subFigSz, height=\subFigH]{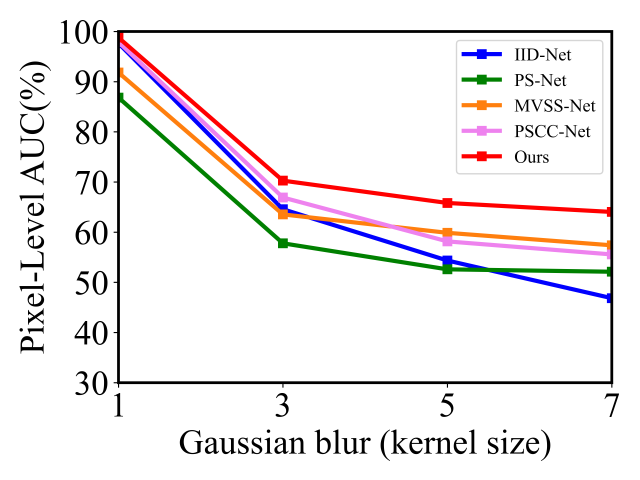}\hspace{0pt}
    \includegraphics[width=\subFigSz, height=\subFigH]{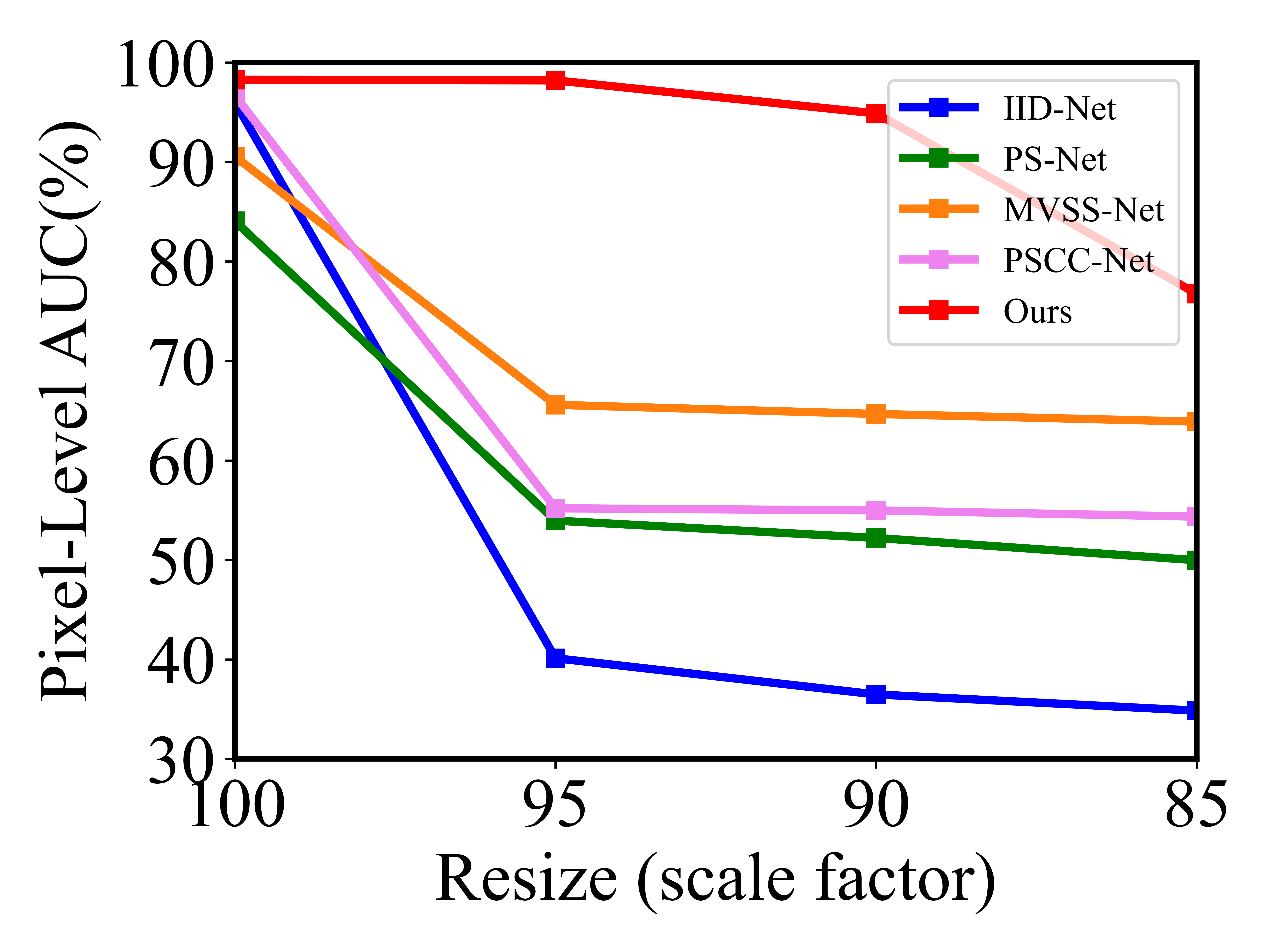}
\end{minipage}

\begin{minipage}[c]{0.99\linewidth}
\captionsetup{justification=centering}
\begin{minipage}[c]{\subCaptionSzA}
    \caption*{{(a)}}
\end{minipage}
% \hspace{-0.01in}
\begin{minipage}[c]{\subCaptionSzB}
    \caption*{{(b)}}
\end{minipage}
\begin{minipage}[c]{\subCaptionSzC}
    \caption*{{(c)}}
\end{minipage}
\begin{minipage}[c]{\subCaptionSzD}
    \caption*{{(d)}}
\end{minipage}
\begin{minipage}[c]{\subCaptionSzE}
    \caption*{{(e)}}
\end{minipage}
\end{minipage}
\vspace{-0.3cm}
\caption{Robustness evaluation on postprocessed IID-9K dataset. Best viewed in color and with zoom-in.}
\vspace{-0.2cm}
\label{fig:roub}
\end{figure*}

\def\subFigSz{0.134\linewidth}
\def\subFigH{0.134\linewidth}
% \captionsetup{type=figure} 
% \includegraphics[]{
\begin{figure*}[t]
\centering
\begin{minipage}[c]{\subFigSz}
     \caption*{\rotatebox{0}{\quad\quad\quad\quad\quad\quad\;\;\scriptsize{{Image}}}}   
\vspace{-0.25cm}
\end{minipage}
\begin{minipage}[c]{\subFigSz}
    \caption*{\rotatebox{0}{\quad\quad\quad\quad\quad\quad\;\;\scriptsize{{GT}}}}
\vspace{-0.25cm}
\end{minipage}
\begin{minipage}[c]{\subFigSz}
    \caption*{\rotatebox{0}{\quad\quad\quad\quad\scriptsize{{PSCC-Net~\cite{liu2022pscc}}}}}
\vspace{-0.25cm}
\end{minipage}
\begin{minipage}[c]{\subFigSz}
    \caption*{\rotatebox{0}{\quad\quad\quad\;\scriptsize{{MVSS-Net  \cite{dong2022mvss}}}}}
\vspace{-0.25cm}
\end{minipage}
\begin{minipage}[c]{\subFigSz}
    \caption*{\rotatebox{0}{\quad\quad\quad\;\scriptsize{{IID-Net \cite{wu2021iid} }}}}
\vspace{-0.25cm}
\end{minipage}
\begin{minipage}[c]{\subFigSz}
    \caption*{\rotatebox{0}{\quad\;\;\scriptsize{{PS-Net \cite{zhang2023ps} }}}}
\vspace{-0.25cm}
\end{minipage}
\begin{minipage}[c]{\subFigSz}
    \caption*{\rotatebox{0}{\scriptsize{{DeFI-Net (Ours)}}}}
\vspace{-0.25cm}
\end{minipage}

\begin{minipage}[t]{0.04\linewidth}
    \raggedleft
    \caption*{\scriptsize{\quad\quad\;\;\;\;GC}}
\end{minipage}\hfill
\begin{minipage}[r]{0.9\linewidth}
    \includegraphics[width=\subFigSz, height=\subFigH]{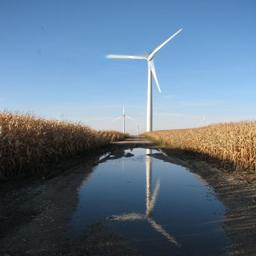}
    \includegraphics[width=\subFigSz, height=\subFigH]{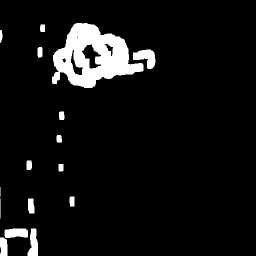}
\includegraphics[width=\subFigSz, height=\subFigH]{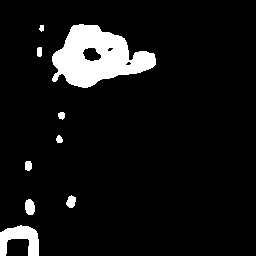}
\includegraphics[width=\subFigSz, height=\subFigH]{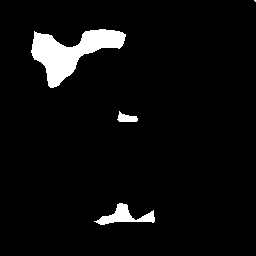}
\includegraphics[width=\subFigSz, height=\subFigH]{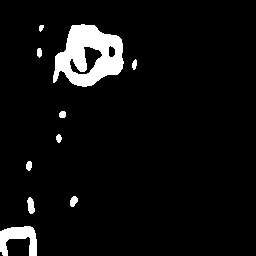}
\includegraphics[width=\subFigSz, height=\subFigH]{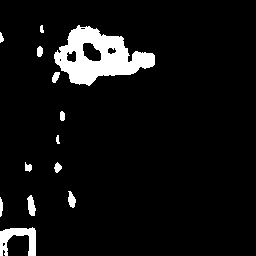}
\includegraphics[width=\subFigSz, height=\subFigH]{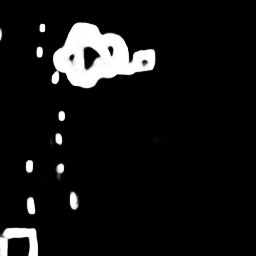}
\end{minipage}

\vspace{0.1cm}

\begin{minipage}[t]{0.04\linewidth}
    \raggedleft
    \caption*{\scriptsize{\quad\quad\;\;\;\;LB}}
\end{minipage}\hfill
\begin{minipage}[r]{0.9\linewidth}
    \includegraphics[width=\subFigSz, height=\subFigH]{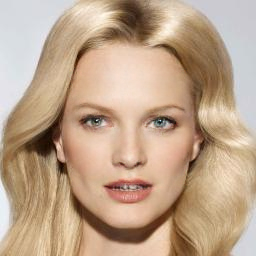}
\includegraphics[width=\subFigSz, height=\subFigH]{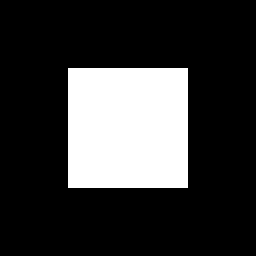}
\includegraphics[width=\subFigSz, height=\subFigH]{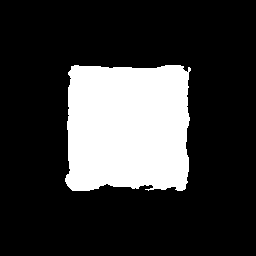}
\includegraphics[width=\subFigSz, height=\subFigH]{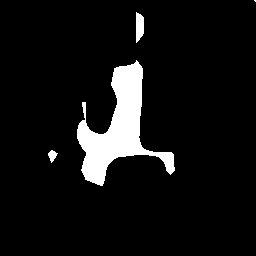}
\includegraphics[width=\subFigSz, height=\subFigH]{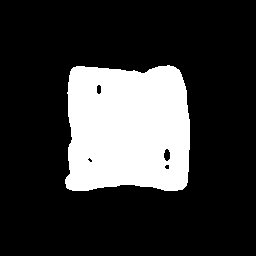}
\includegraphics[width=\subFigSz, height=\subFigH]{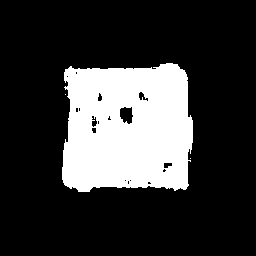}
\includegraphics[width=\subFigSz, height=\subFigH]{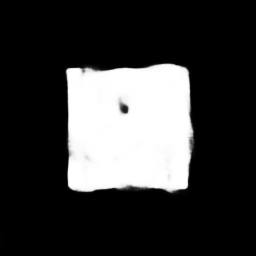}
\end{minipage}

\vspace{0.1cm}

\begin{minipage}[t]{0.02\linewidth}
    \raggedleft
    \caption*{\scriptsize{\quad\quad\;\;\;\;NS}}
\end{minipage}\hfill
\begin{minipage}[r]{0.9\linewidth}
  \includegraphics[width=\subFigSz, height=\subFigH]{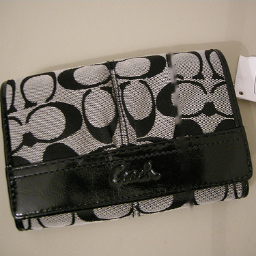}
\includegraphics[width=\subFigSz, height=\subFigH]{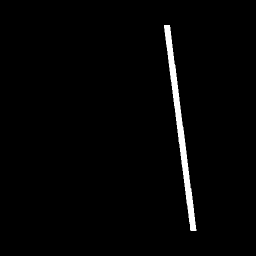}
\includegraphics[width=\subFigSz, height=\subFigH]{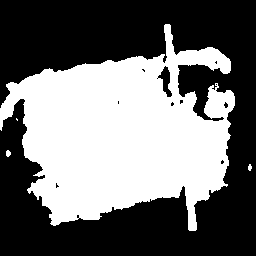}
\includegraphics[width=\subFigSz, height=\subFigH]{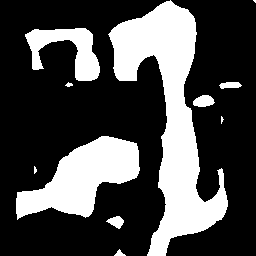}
\includegraphics[width=\subFigSz, height=\subFigH]{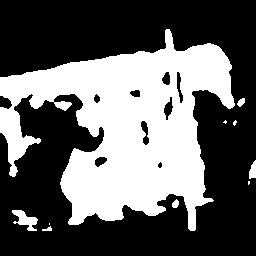}
\includegraphics[width=\subFigSz, height=\subFigH]{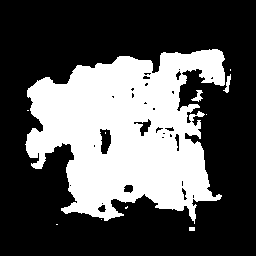}
\includegraphics[width=\subFigSz, height=\subFigH]{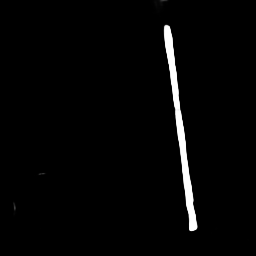}
\end{minipage}

\vspace{0.1cm}

\begin{minipage}[t]{0.04\linewidth}
    \raggedleft
    \caption*{\scriptsize{\quad\;\;DEFACTO}}
\end{minipage}\hfill
\begin{minipage}[r]{0.9\linewidth}
\includegraphics[width=\subFigSz, height=\subFigH]{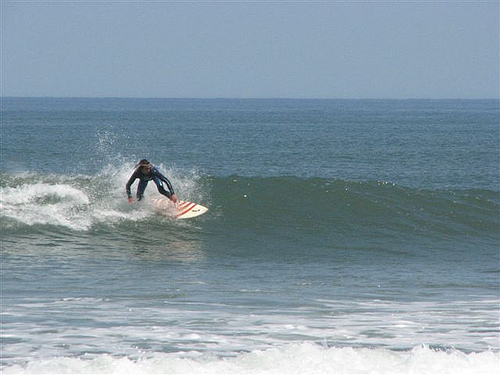}
\includegraphics[width=\subFigSz, height=\subFigH]{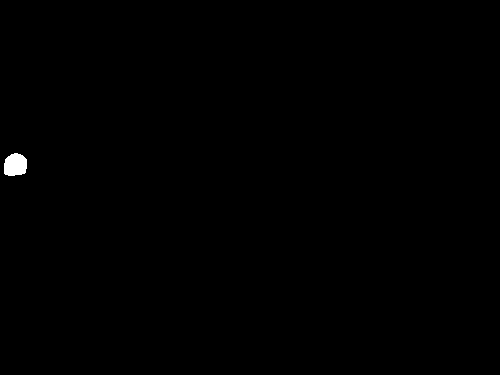}
\includegraphics[width=\subFigSz, height=\subFigH]{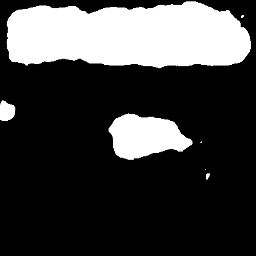}
\includegraphics[width=\subFigSz, height=\subFigH]{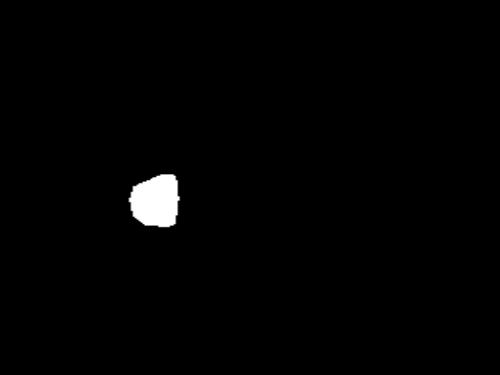}
\includegraphics[width=\subFigSz, height=\subFigH]{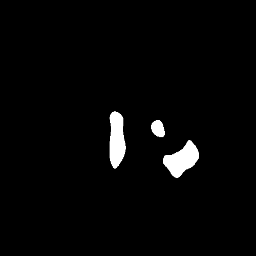}
\includegraphics[width=\subFigSz, height=\subFigH]{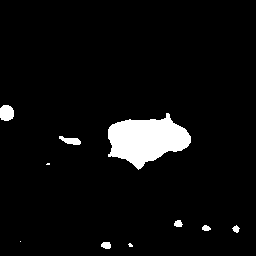}
\includegraphics[width=\subFigSz, height=\subFigH]{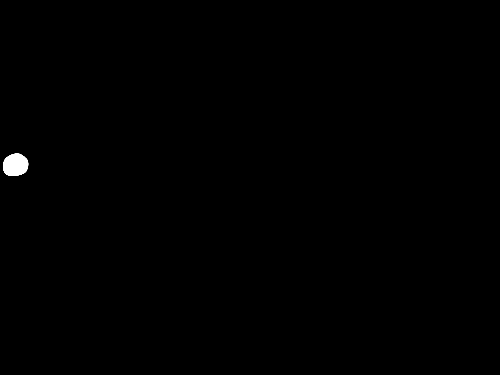}
\end{minipage}

\vspace{0.1cm}

\begin{minipage}[t]{0.04\linewidth}
    \raggedleft
    \caption*{\scriptsize{\quad\;\;Photoshop}}
\end{minipage}\hfill
\begin{minipage}[r]{0.9\linewidth}
 \includegraphics[width=\subFigSz, height=\subFigH]{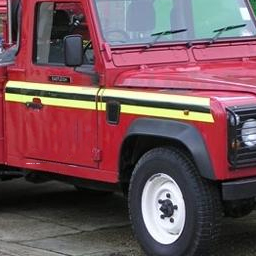}
\includegraphics[width=\subFigSz, height=\subFigH]{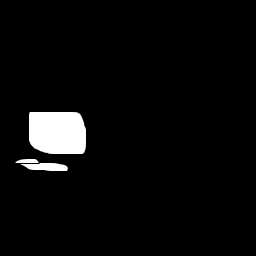}
\includegraphics[width=\subFigSz, height=\subFigH]{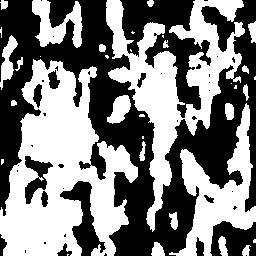}
\includegraphics[width=\subFigSz, height=\subFigH]{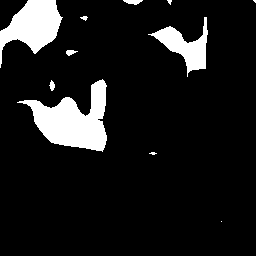}
\includegraphics[width=\subFigSz, height=\subFigH]{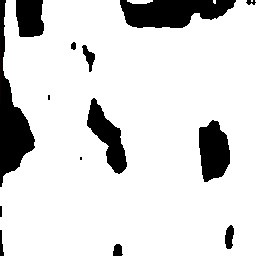}
\includegraphics[width=\subFigSz, height=\subFigH]{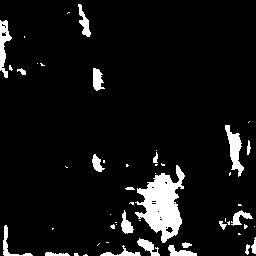}
\includegraphics[width=\subFigSz, height=\subFigH]{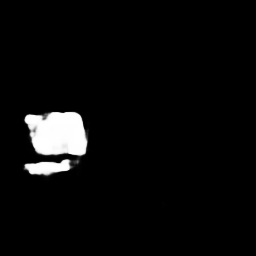}
\end{minipage}

\vspace{0.1cm}

\begin{minipage}[r]{0.04\linewidth}
    \raggedleft
     \caption*{\scriptsize{\quad\;\;AutoSplice}}
\end{minipage}\hfill
\begin{minipage}[r]{0.9\linewidth}
\includegraphics[width=\subFigSz, height=\subFigH]{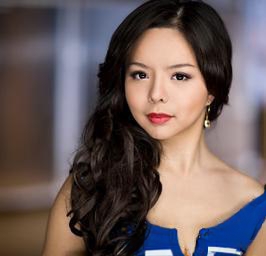}
\includegraphics[width=\subFigSz, height=\subFigH]{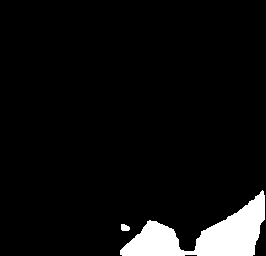}
\includegraphics[width=\subFigSz, height=\subFigH]{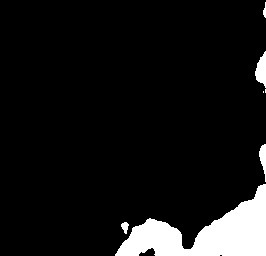}
\includegraphics[width=\subFigSz, height=\subFigH]{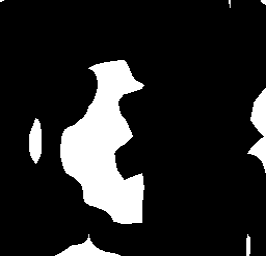}
\includegraphics[width=\subFigSz, height=\subFigH]{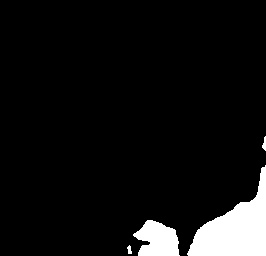}
\includegraphics[width=\subFigSz, height=\subFigH]{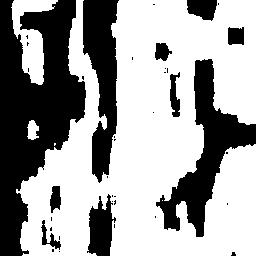}
\includegraphics[width=\subFigSz, height=\subFigH]{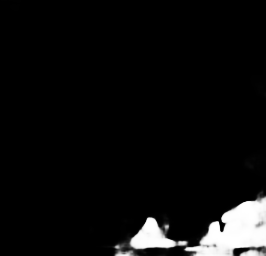}
\end{minipage}
\caption{{Qualitative comparisons for localizing forgeries from various inpainting models, including in-domain testing results on the GC data from~\cite{wu2021iid}, and cross-domain testing results on the LB and NS data from~\cite{wu2021iid}, DEFACTO~\cite{mahfoudi2019defacto}, Photoshop~\cite{zhang2023ps}, and AutoSplice~\cite{jia2023autosplice} datasets.}} 
\vspace{-0.4cm}
\label{fig:qua}
\end{figure*}

\subsection{Qualitative Comparisons} 

{In Fig.~\ref{fig:qua}, we demonstrate the qualitative results on several image inpainting datasets. Our method achieves outstanding accuracy in localizing image inpainting across diverse manipulation regions and models, significantly surpassing the baseline methods. The segmentation results produced by DeFI-Net are closest to the ground truth, with rare occurrences of false positives. Specifically, DeFI-Net exhibits excellent performance in handling random regions (first and third rows) and specific semantic objects (second, fourth, fifth, and sixth rows) for inpainting. In contrast, MVSS-Net, IID-Net, and PS-Net produce unsatisfactory results, with many missegmented regions and even overlooking the correct targets. PSCC-Net achieves decent results on some datasets, but it is prone to misjudgment in some challenging cases (such as slender, small targets) due to the influence of semantic information. In response to this issue, DeFI-Net obtains precise predictions thanks to our proposed dense feature pyramid learning strategy. This strategy explores the potential relationships among multi-level features, better learning the inconsistencies in image details, structure, and semantics, enabling the model to better handle inpainting at different scales. Moreover, the reverse edge attention enhancement effectively enhances edge features and suppresses the impact of semantic noise on inpainting representation, resulting in clearer boundary localization. Overall, DeFI-Net achieves satisfactory results.}

In addition, we found that DeFI-Net significantly improved the localization of small target inpainting compared to existing methods. Removing unnecessary people or objects in the background is the most commonly used application scenario for image inpainting. Tiny inpainting makes minor changes to the semantics of the image and leaves behind artifacts that are difficult to detect, which greatly increases the localization difficulty. As shown in Fig. \ref{fig:tiny}, IID-Net and MVSS-Net have failed, and PS-Net and PSCC-Net have many false positives, while our method achieves accurate localization. This primarily benefits from our DFPL and SA-FF, which dig deep into the correlation between multi-scale information and skillfully combine them.
%It is worth noting that DE-24K contains authentic images, and DeFI-Net, which is pre-trained in it, can also make correct predictions when confronted with uninpainted samples.} 

\subsection{{Computational Complexity Comparisons}}

To comprehensively evaluate the proposed method, we present comparisons of time consumption and computational complexity in Table \ref{tab:t1}. All the compared methods are evaluated on the same device (GeForce RTX 3090 GPUs) with $256 \times 256$ images. We can see that the proposed DeFI-Net introduces little computation overhead, and achieves comparable performance to existing models in terms of inference time.

\begin{table}[]
\centering
\caption{Comparison of computational complexity and time consumption. The run time is averaged over 100 images.}
\newcommand{\tabincell}[2]{\begin{tabular}{@{}#1@{}}#2\end{tabular}} 
\scalebox{1.1}{\begin{tabular}{r|c|c|c}
\Xhline{0.8pt}
Model$\;$  & \tabincell{c}{Flops\\(GFlops)} & \tabincell{c}{Parameters \\($10^6$)} & \tabincell{c}{Run time\\ (sec/image)} \\ \hline
IID-Net \cite{wu2021iid}     & 32.93         & 5.79   &  0.055  \\  %5787567
PSCC-Net \cite{liu2022pscc}     & 26.62         & 3.67    &  0.022 \\  %3667942
MVSS-Net \cite{dong2022mvss}     & 40.84         & 146.81  &  0.017 \\  %\hline % 146811215
PS-Net \cite{zhang2023ps}    & -              & -       & 0.024 \\ \hline
HRNet \cite{wang2020deep} &  24.37     & 2.02   &  - \\
DeFI-Net      & 43.00         &  2.82   &  0.025 \\   % 32669836
\Xhline{0.8pt}
%Res2Net \cite{gao2019res2net}    &  11.81        &  25.72  &  - \\
%DeFI-Net (Res2Net)    & 14.46     & 32.67   &  0.022 \\   \hline
%Our(c=256) & 62.59         & 64387682   \\ \hline
\end{tabular}}
\vspace{-0.2cm}
\label{tab:t1}
\end{table}

\def\subFigSz{0.12\linewidth}
\def\subFigH{0.085\linewidth}
% \captionsetup{type=figure} 
% \includegraphics[]{
\begin{figure*}[t]
\centering
\begin{minipage}[c]{0.05\linewidth}
     \caption*{\rotatebox{0}{\scriptsize{{Authentic}}}}   
\vspace{-0.25cm}
\end{minipage}
\begin{minipage}[c]{\subFigSz}
     \caption*{\rotatebox{0}{\quad\quad\quad\;\;\scriptsize{{Inpainting}}}}   
\vspace{-0.25cm}
\end{minipage}
\begin{minipage}[c]{\subFigSz}
    \caption*{\rotatebox{0}{\quad\quad\quad\;\;\scriptsize{{GT}}}}
\vspace{-0.25cm}
\end{minipage}
\begin{minipage}[c]{\subFigSz}
    \caption*{\rotatebox{0}{\quad\quad\quad\scriptsize{{PSCC-Net~\cite{liu2022pscc}}}}}
\vspace{-0.25cm}
\end{minipage}
\begin{minipage}[c]{\subFigSz}
    \caption*{\rotatebox{0}{\quad\quad\;\;\scriptsize{{MVSS-Net  \cite{dong2022mvss}}}}}
\vspace{-0.25cm}
\end{minipage}
\begin{minipage}[c]{\subFigSz}
    \caption*{\rotatebox{0}{\quad\quad\quad\;\scriptsize{{IID-Net \cite{wu2021iid} }}}}
\vspace{-0.25cm}
\end{minipage}
\begin{minipage}[c]{\subFigSz}
    \caption*{\rotatebox{0}{\quad\quad\;\;\;\scriptsize{{PS-Net \cite{zhang2023ps} }}}}
\vspace{-0.25cm}
\end{minipage}
\begin{minipage}[c]{\subFigSz}
    \caption*{\rotatebox{0}{\quad\quad\scriptsize{{DeFI-Net (Ours)}}}}
\vspace{-0.25cm}
\end{minipage}

% \begin{minipage}[t]{0.04\linewidth}
%     \raggedleft
%     \caption*{\scriptsize{\quad\quad\;\;\;\;GC}}
% \end{minipage}\hfill
\begin{minipage}[r]{\linewidth}
\includegraphics[width=\subFigSz, height=\subFigH]{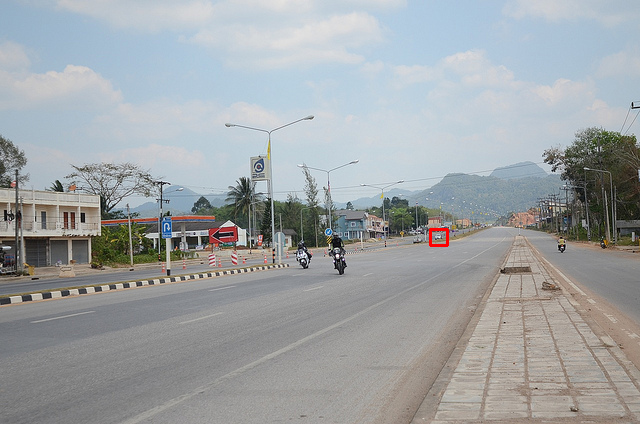} 
\includegraphics[width=\subFigSz, height=\subFigH]{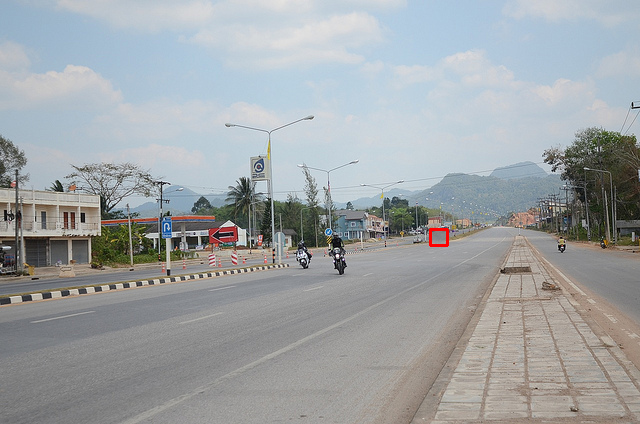} 
\includegraphics[width=\subFigSz, height=\subFigH]{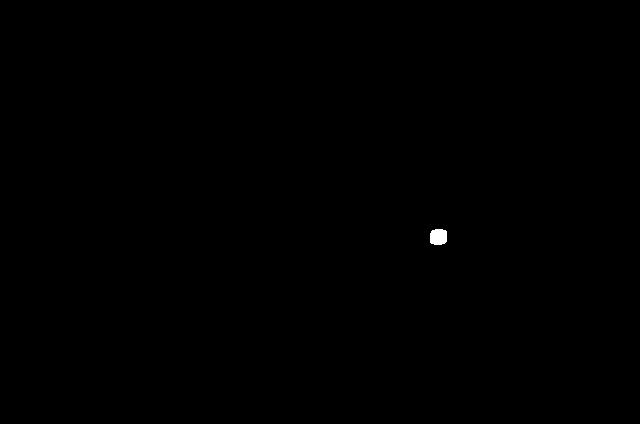} 
\includegraphics[width=\subFigSz, height=\subFigH]{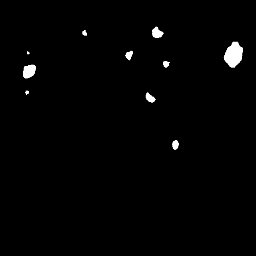} 
\includegraphics[width=\subFigSz, height=\subFigH]{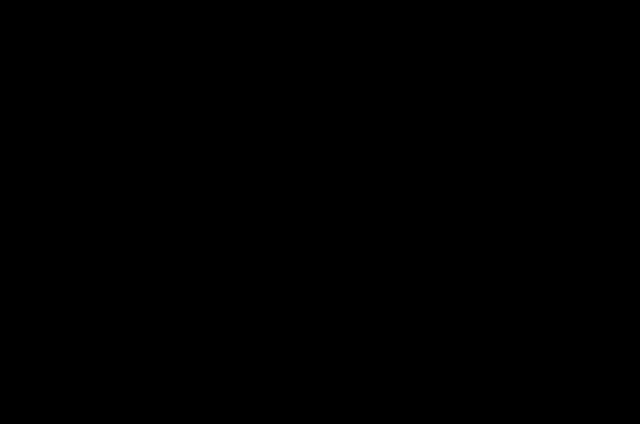} 
\includegraphics[width=\subFigSz, height=\subFigH]{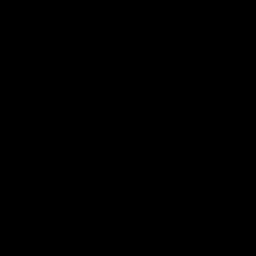} 
\includegraphics[width=\subFigSz, height=\subFigH]{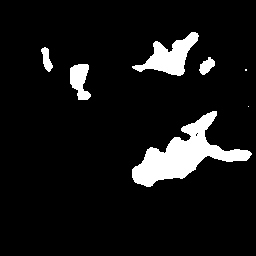} 
\includegraphics[width=\subFigSz, height=\subFigH]{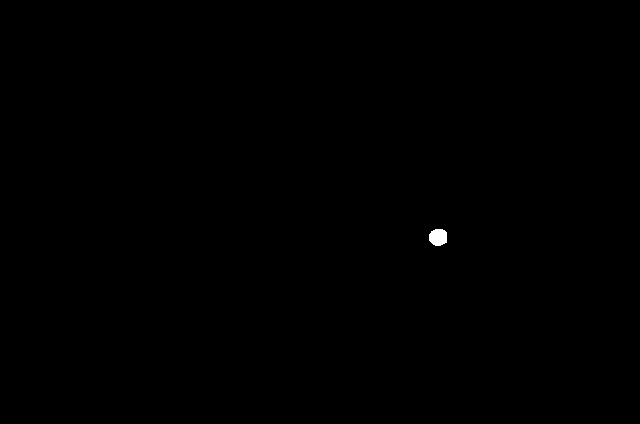} 
\end{minipage}

\vspace{0.1cm}

% \begin{minipage}[t]{0.04\linewidth}
%     \raggedleft
%     \caption*{\scriptsize{\quad\quad\;\;\;\;LB}}
% \end{minipage}\hfill
\begin{minipage}[r]{\linewidth}
\includegraphics[width=\subFigSz, height=\subFigH]{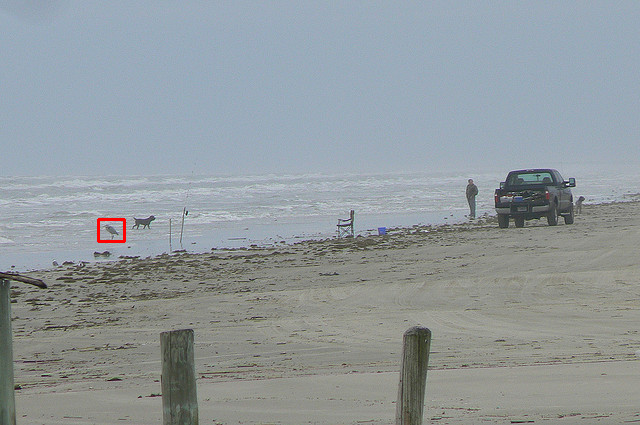}
\includegraphics[width=\subFigSz, height=\subFigH]{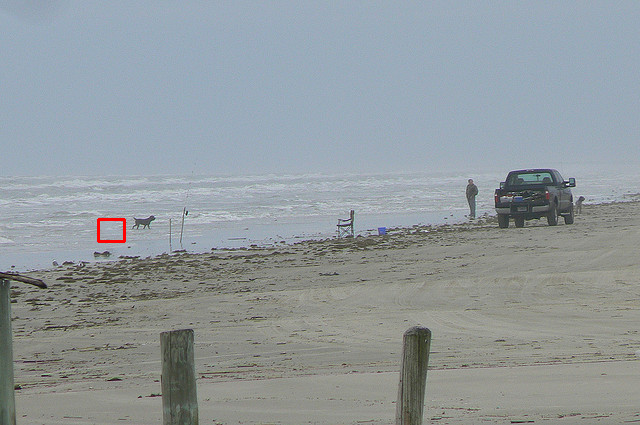}
\includegraphics[width=\subFigSz, height=\subFigH]{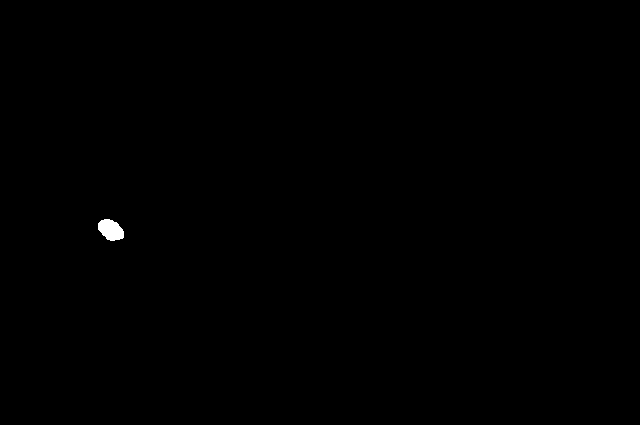}
\includegraphics[width=\subFigSz, height=\subFigH]{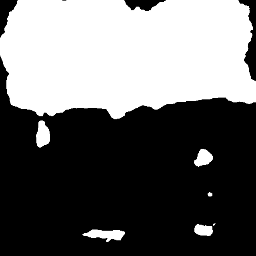}
\includegraphics[width=\subFigSz, height=\subFigH]{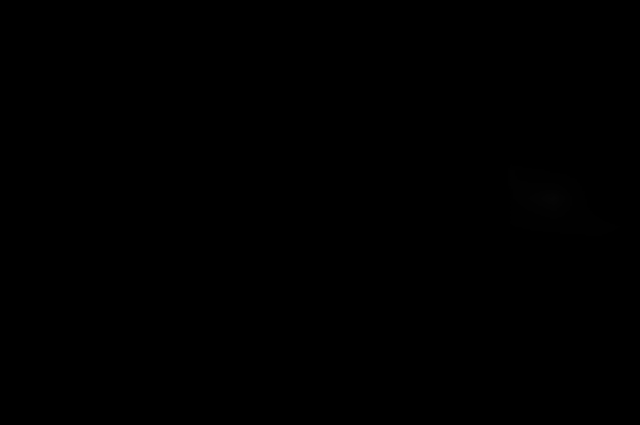}
\includegraphics[width=\subFigSz, height=\subFigH]{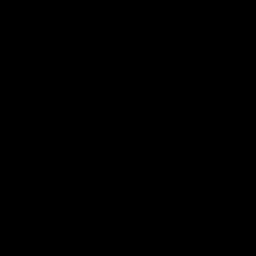}
\includegraphics[width=\subFigSz, height=\subFigH]{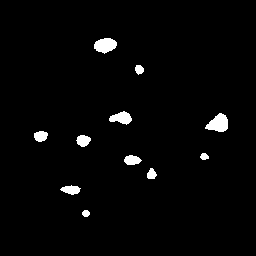}
\includegraphics[width=\subFigSz, height=\subFigH]{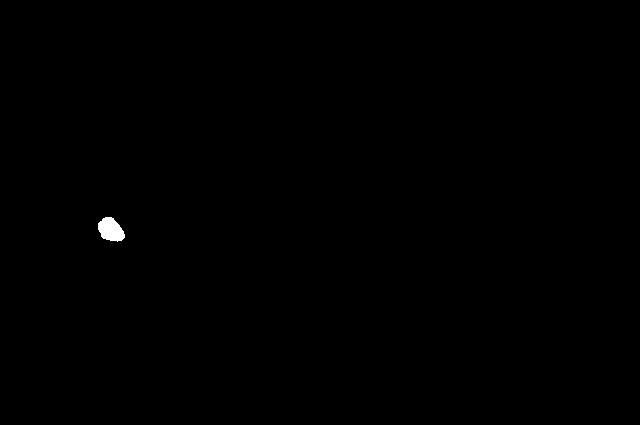}
\end{minipage}

\vspace{0.1cm}

% \begin{minipage}[t]{0.02\linewidth}
%     \raggedleft
%     \caption*{\scriptsize{\quad\quad\;\;\;\;NS}}
% \end{minipage}\hfill
\begin{minipage}[r]{\linewidth}
\includegraphics[width=\subFigSz, height=\subFigH]{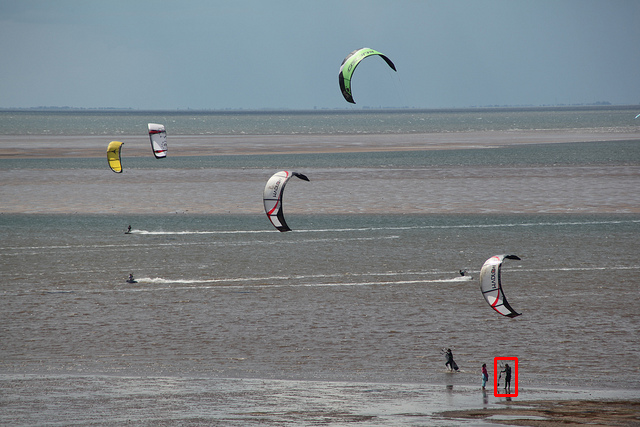}
\includegraphics[width=\subFigSz, height=\subFigH]{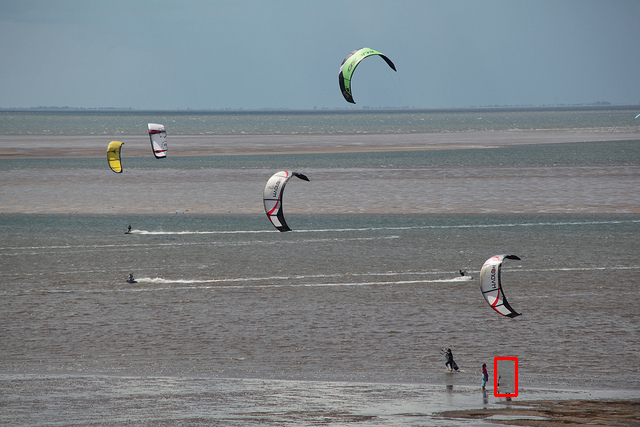}
\includegraphics[width=\subFigSz, height=\subFigH]{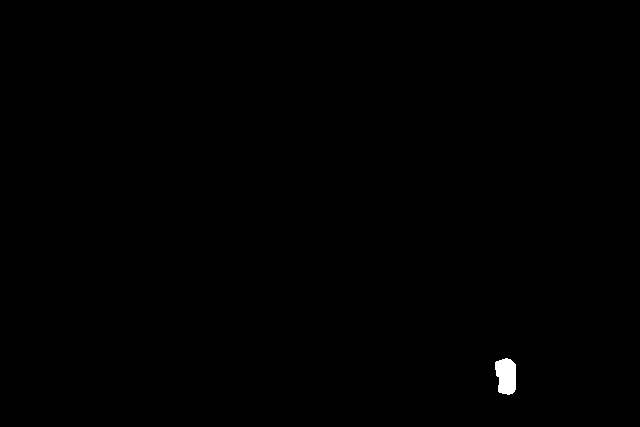}
\includegraphics[width=\subFigSz, height=\subFigH]{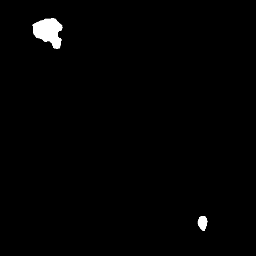}
\includegraphics[width=\subFigSz, height=\subFigH]{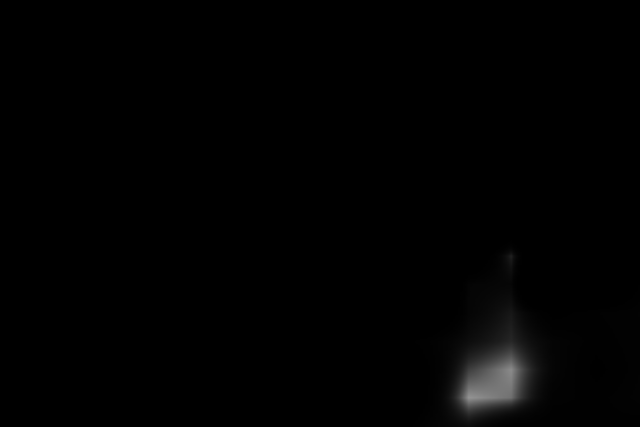}
\includegraphics[width=\subFigSz, height=\subFigH]{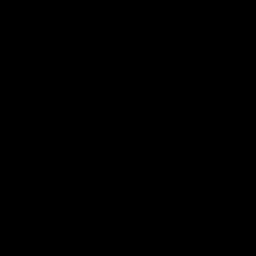}
\includegraphics[width=\subFigSz, height=\subFigH]{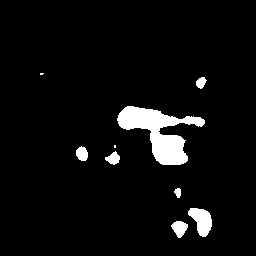}
\includegraphics[width=\subFigSz, height=\subFigH]{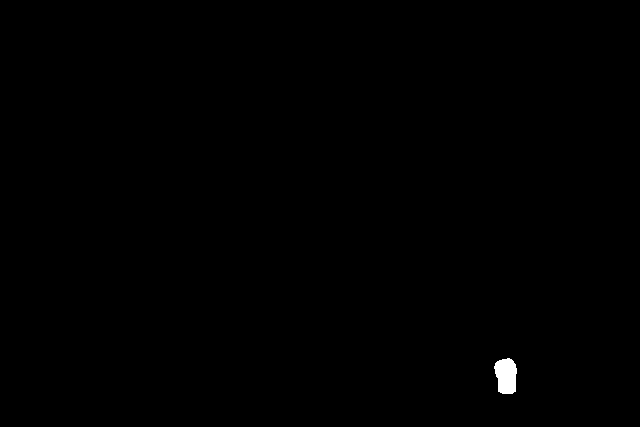}
\end{minipage}

\vspace{0.1cm}

% \begin{minipage}[t]{0.04\linewidth}
%     \raggedleft
%     \caption*{\scriptsize{\quad\;\;DEFACTO}}
% \end{minipage}\hfill
\begin{minipage}[r]{\linewidth}
\includegraphics[width=\subFigSz, height=\subFigH]{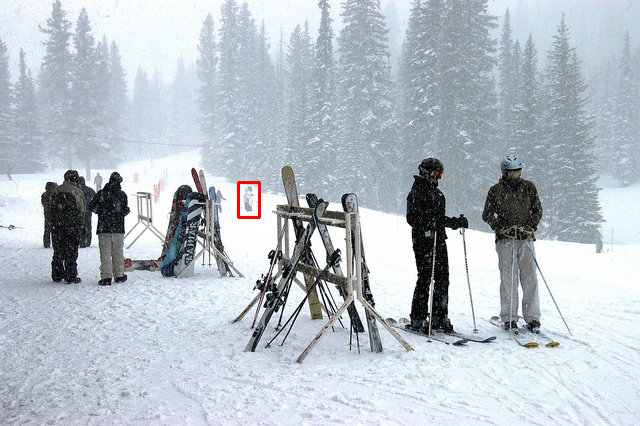}
\includegraphics[width=\subFigSz, height=\subFigH]{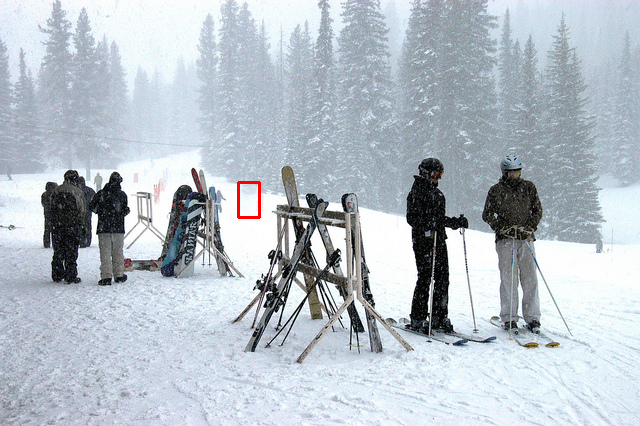}
\includegraphics[width=\subFigSz, height=\subFigH]{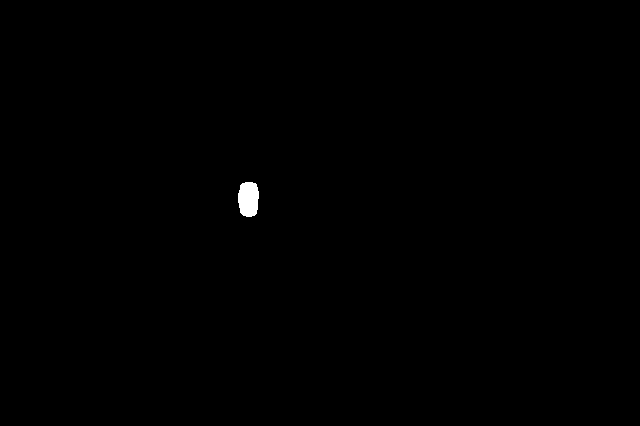}
\includegraphics[width=\subFigSz, height=\subFigH]{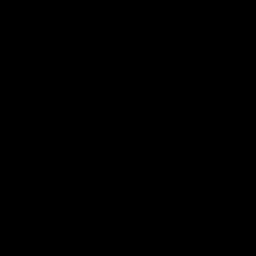}
\includegraphics[width=\subFigSz, height=\subFigH]{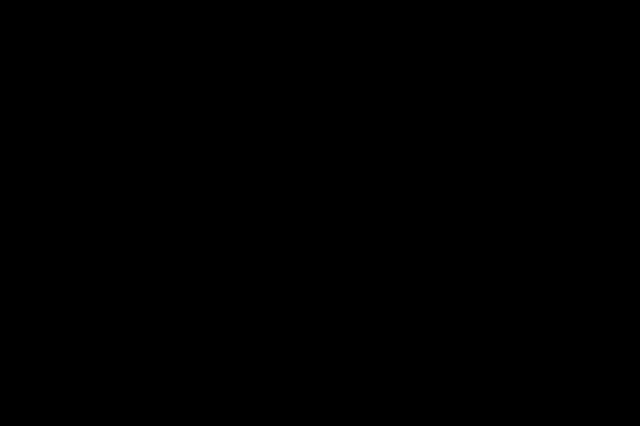}
\includegraphics[width=\subFigSz, height=\subFigH]{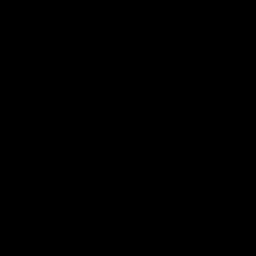}
\includegraphics[width=\subFigSz, height=\subFigH]{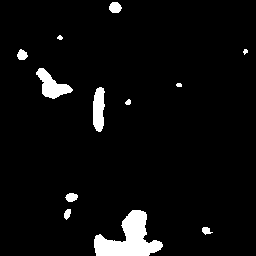}
\includegraphics[width=\subFigSz, height=\subFigH]{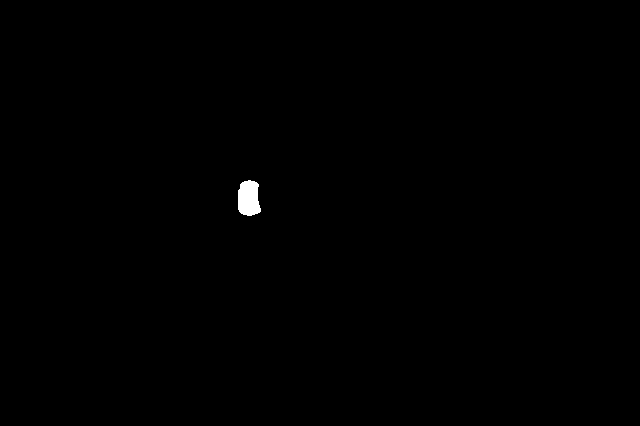}
\end{minipage}

\vspace{0.1cm}

% \begin{minipage}[t]{0.04\linewidth}
%     \raggedleft
%     \caption*{\scriptsize{\quad\;\;Photoshop}}
% \end{minipage}\hfill
\begin{minipage}[r]{\linewidth}
\includegraphics[width=\subFigSz, height=\subFigH]{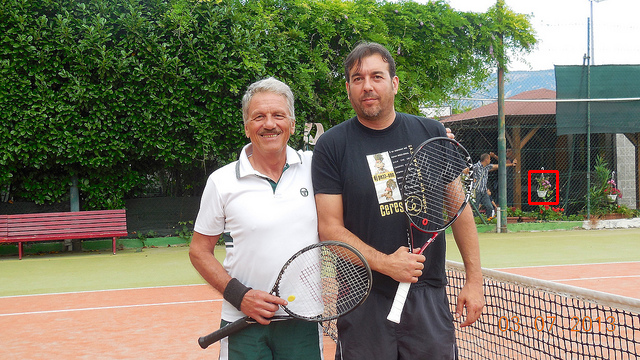} 
\includegraphics[width=\subFigSz, height=\subFigH]{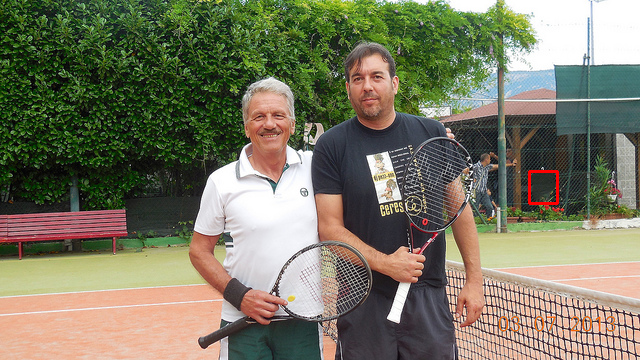}
\includegraphics[width=\subFigSz, height=\subFigH]{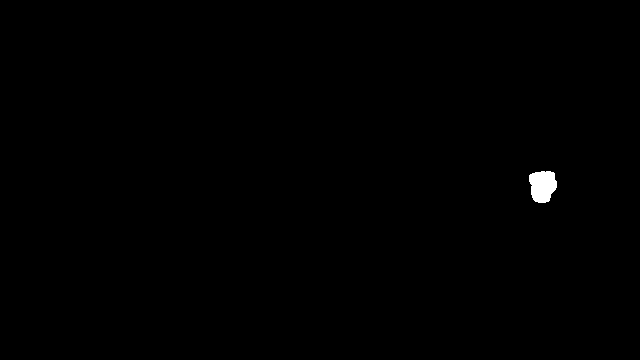}
\includegraphics[width=\subFigSz, height=\subFigH]{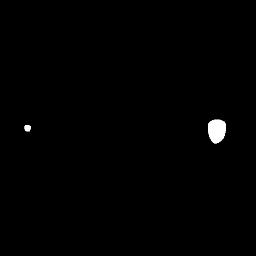}
\includegraphics[width=\subFigSz, height=\subFigH]{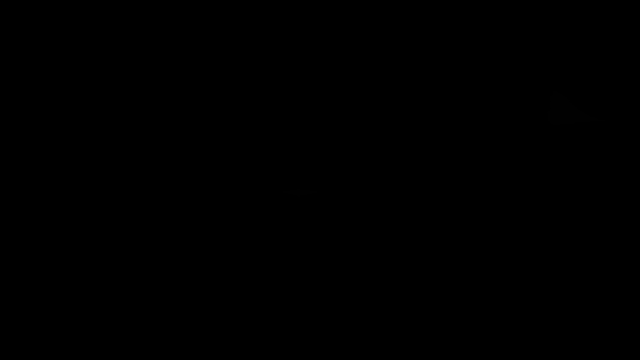}
\includegraphics[width=\subFigSz, height=\subFigH]{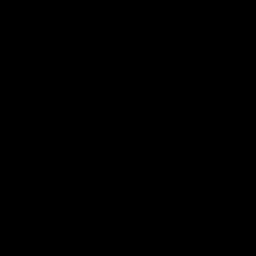}
\includegraphics[width=\subFigSz, height=\subFigH]{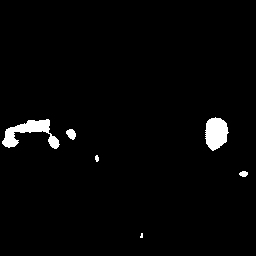}
\includegraphics[width=\subFigSz, height=\subFigH]{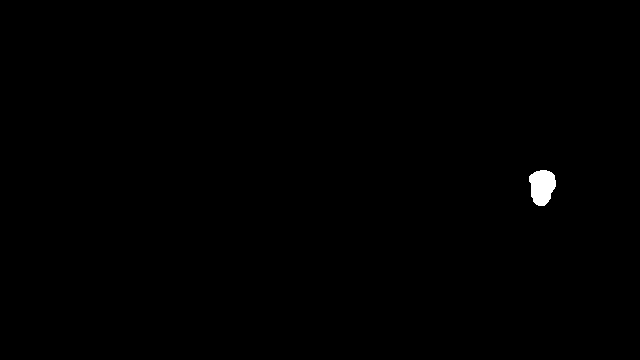}
\end{minipage}

\caption{{Qualitative comparison of locating tiny inpainting on the DEFACTO~\cite{mahfoudi2019defacto} dataset. Our method accurately predicts inpainting areas, whereas IID-Net and MVSS-Net have failed, PS-Net and PSCC-Net suffer from many false positives.}}
\vspace{-0.4cm}
\label{fig:tiny}
\end{figure*}

\subsection{{Visualization Results}} 
{To further present the discriminative and complementary features learned from various levels by DeFI-Net, we provide the visualization results of the feature maps at the low, middle, and high levels. The intermediate features to be observed were iterated along the channel dimension and converted into PIL images using the \textit{torchvision.transforms.ToPILImage()}. The results are shown in Fig.~\ref{fig:visual}. We can observe that the feature maps progressively capture the required discriminative information from low-level to high-level, including details, local structures, and global abstract semantics. Low-level features capture discriminative patterns around edges, middle-level features focus on texture inconsistencies in the filled areas, while high-level features identify changes in an image’s semantic content, such as localizing the entire objects. By effectively integrating these features, our model achieves precise localization across various inpainting regions and generation models.}

\subsection{Ablation Studies}
{Essentially, DeFI-Net mainly benefits from three modules: DFPL, REAE, and SA-FF. Dense Feature Pyramid Learning (DFPL) highlights the specificity of features at different stages in the neural network, interacts quickly with neighboring layers in a densely connected manner, and improves the representation capability of features from other perspectives. Reverse Edge Attention Enhancement (REAE) enables the model to capture subtle differences at the boundaries by utilizing structural information provided by mid-level features as a priori guide. Spatial Adaptive Feature Fusion (SA-FF) adaptively learns how high-level features are fused with low and mid-level features. To evaluate the effectiveness of DFPL, REAE, and SA-FF, we removed them separately from DeFI-Net and evaluated the inpainting detection performance on the IID-10K dataset with the ablation setup shown in Table~\ref{tab:4}. In addition, we performed ablation analysis for each module separately to demonstrate the effectiveness of their design.}

\noindent\textbf{Effect of DFPL.} Previous works typically use either top-down \cite{lin2017feature} or bottom-up \cite{liu2018path} for information interaction among features of different stages. While these methods alleviate the inefficiency of propagating top-level or bottom-level information in the opposite direction, they neglect the specificity of multilevel features. On the other hand, the densely connected method can fully utilize the adjacent and cross-layer information to enhance the context-awareness of features at each level. As can be observed from Table~\ref{tab:5}, compared with Var\#1 (FPN \cite{lin2017feature}) and Var\#2 (PANet \cite{liu2018path}), DFPL significantly enhances performance and effectively promotes the spreading and utilization of multi-level features. An interesting phenomenon can be observed: the performance of DFPL with concatenation fusion slightly outperforms that with addition fusion. This difference may arise because the addition method tends to suppress detailed information, whereas dimension reduction with a $1\times1$ convolution after concatenation serves as a feature selection process, aiding in the interaction of low-level features.
\def\subFigSz{0.155\linewidth}
\def\subFigH{0.155\linewidth}
% \captionsetup{type=figure} 
% \includegraphics[]{
\begin{figure}[t]
\centering
\begin{minipage}[c]{\subFigSz}
    \caption*{\scriptsize{\quad\quad\quad\quad\;\;\;Image}}
\vspace{-0.25cm}
\end{minipage}
\begin{minipage}[c]{\subFigSz}
    \caption*{\scriptsize{\quad\quad\quad\quad\;\;\;GT}}
\vspace{-0.25cm}
\end{minipage}
\begin{minipage}[c]{\subFigSz}
    \caption*{\scriptsize{\quad\quad\quad\;\;\;Low}}
\vspace{-0.25cm}
\end{minipage}
\begin{minipage}[c]{\subFigSz}
    \caption*{\scriptsize{\quad\quad\quad\;\;Mid }}
\vspace{-0.25cm}
\end{minipage}
\begin{minipage}[c]{\subFigSz}
    \caption*{\scriptsize{\quad\quad High }}
\vspace{-0.25cm}
\end{minipage}
\begin{minipage}[c]{\subFigSz}
    \caption*{\scriptsize{\;\; Prediction}}
\vspace{-0.25cm}
\end{minipage}

\begin{minipage}[t]{0.04\linewidth}
    \raggedleft
    \caption*{\rotatebox{0}{\tiny{{AutoSplice}}}}
\end{minipage}\hfill
\begin{minipage}[r]{0.9\linewidth}
    \includegraphics[width=\subFigSz, height=\subFigH]{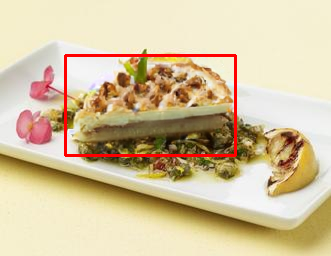}
    \includegraphics[width=\subFigSz, height=\subFigH]{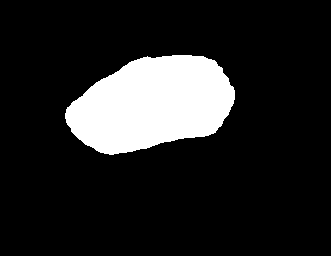}
    \includegraphics[width=\subFigSz, height=\subFigH]{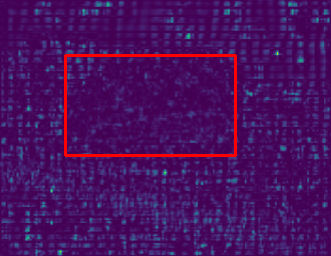}
    \includegraphics[width=\subFigSz, height=\subFigH]{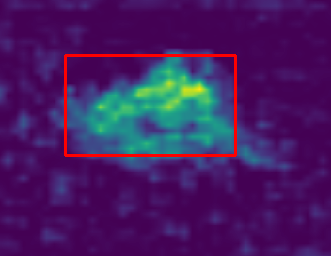}
    \includegraphics[width=\subFigSz, height=\subFigH]{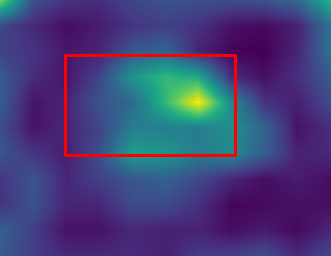}
    \includegraphics[width=\subFigSz, height=\subFigH]{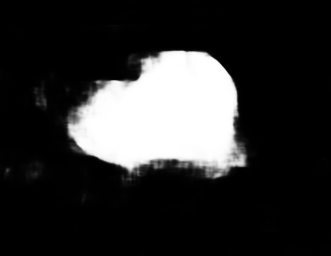}
\end{minipage}

\begin{minipage}[t]{0.04\linewidth}
    \raggedleft
    \caption*{\rotatebox{0}{\;\;\tiny{{IID}}}}
\end{minipage}\hfill
\begin{minipage}[r]{0.9\linewidth}
    \includegraphics[width=\subFigSz, height=\subFigH]{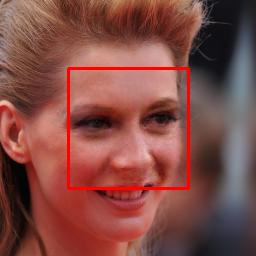}
    \includegraphics[width=\subFigSz, height=\subFigH]{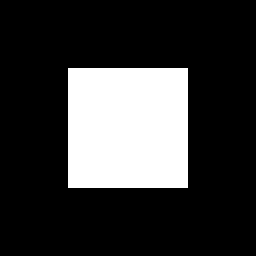}
    \includegraphics[width=\subFigSz, height=\subFigH]{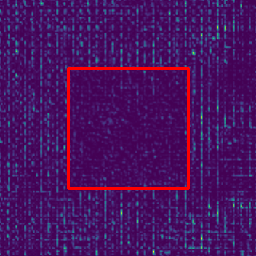}
    \includegraphics[width=\subFigSz, height=\subFigH]{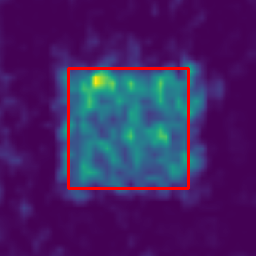}
    \includegraphics[width=\subFigSz, height=\subFigH]{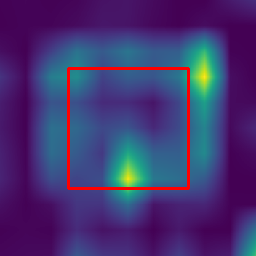}
    \includegraphics[width=\subFigSz, height=\subFigH]{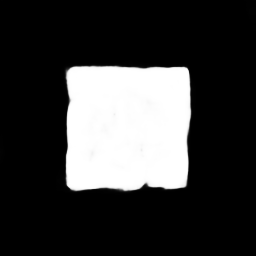}
\end{minipage}

\begin{minipage}[t]{0.04\linewidth}
    \raggedleft
    \caption*{\rotatebox{0}{\tiny{{Photoshop}}}}
\end{minipage}\hfill
\begin{minipage}[r]{0.9\linewidth}
    \includegraphics[width=\subFigSz, height=\subFigH]{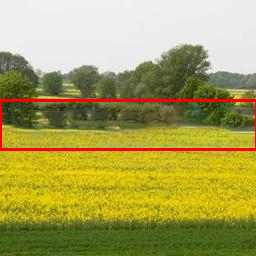}
    \includegraphics[width=\subFigSz, height=\subFigH]{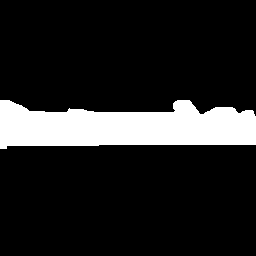}
    \includegraphics[width=\subFigSz, height=\subFigH]{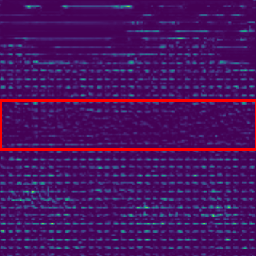}
    \includegraphics[width=\subFigSz, height=\subFigH]{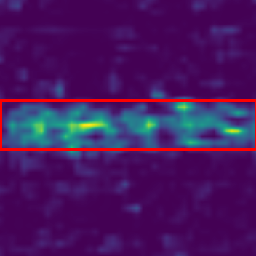}
    \includegraphics[width=\subFigSz, height=\subFigH]{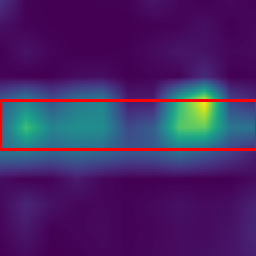}
    \includegraphics[width=\subFigSz, height=\subFigH]{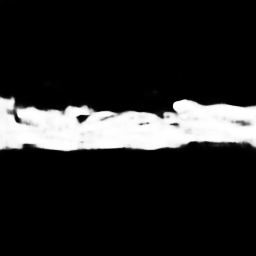}
\end{minipage}

\begin{minipage}[t]{0.04\linewidth}
    \raggedleft
    \caption*{\rotatebox{0}{\tiny{{DEFACTO}}}}
\end{minipage}\hfill
\begin{minipage}[r]{0.9\linewidth}
    \includegraphics[width=\subFigSz, height=\subFigH]{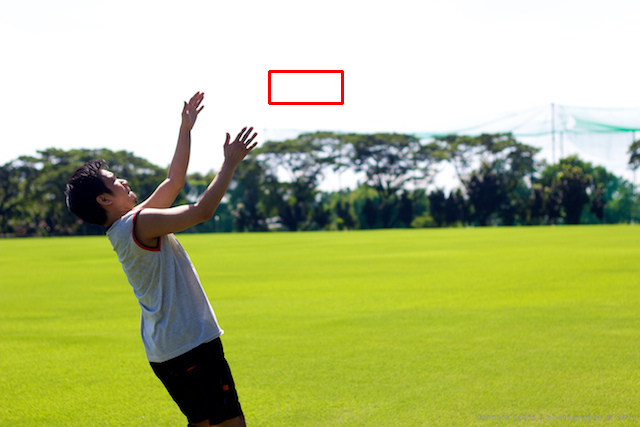}
    \includegraphics[width=\subFigSz, height=\subFigH]{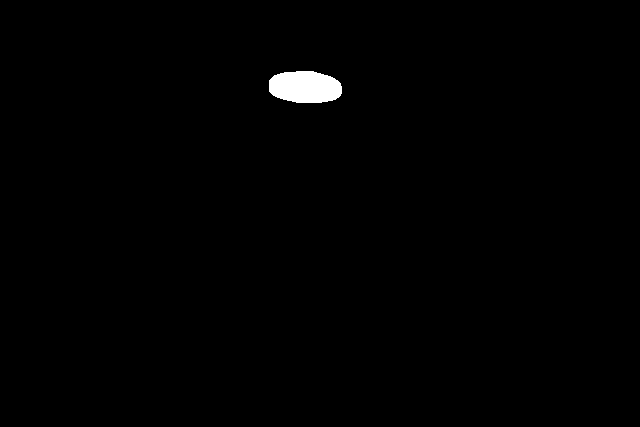}
    \includegraphics[width=\subFigSz, height=\subFigH]{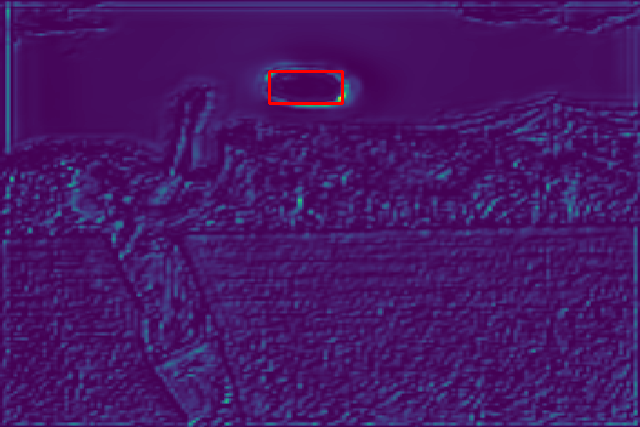}
    \includegraphics[width=\subFigSz, height=\subFigH]{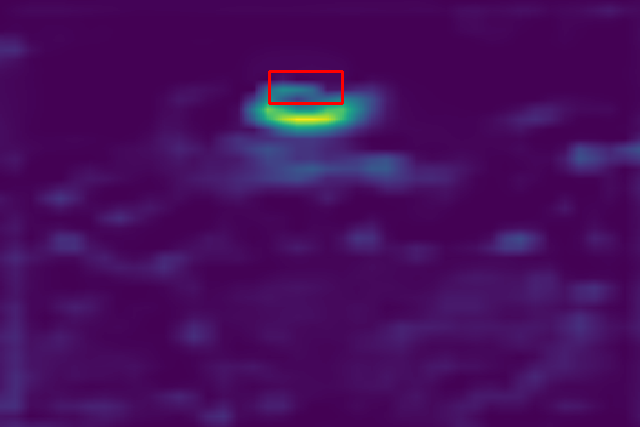}
    \includegraphics[width=\subFigSz, height=\subFigH]{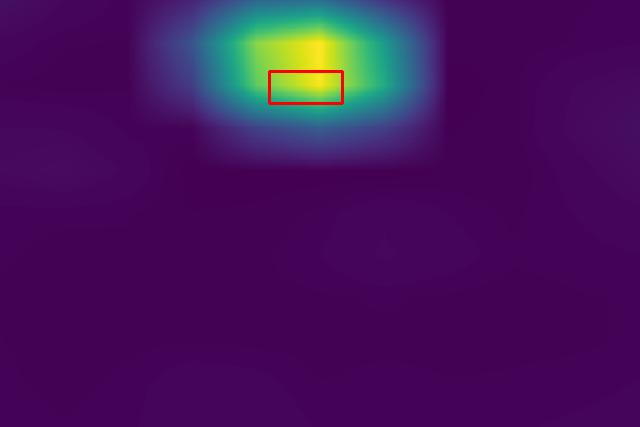}
    \includegraphics[width=\subFigSz, height=\subFigH]{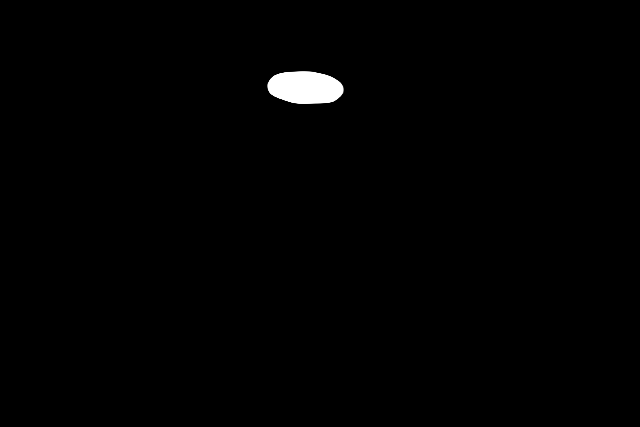}
\end{minipage}
\caption{Visualization of DeFI-Net on various testing datasets. }
\vspace{-0.2cm}
\label{fig:visual}
\end{figure}

\begin{table}[tbp]
\centering
\caption{$F1$ and $AUC$ (\%) of different combinations for our method on IID-10K dataset.}
\label{tab:4}
\setlength{\tabcolsep}{4mm}{
\begin{tabular}{r|l|ll}
\hline
Setup     & Components                             & AUC     &   F1  \\ \hline
Var\#0    & Original HRNet \cite{wang2020deep}      & 92.50   & 74.88 \\
Var\#1    & HRNet+DFPL                           & 97.24   & 85.39 \\
Var\#2    & HRNet+DFPL+REAE                        & 98.25   & 89.40 \\
Var\#3    & HRNet+DFPL+SA-FF                       & 98.15   & 88.14 \\
DeFI-Net   & HRNet+DFPL+REAE+SA-FF                  & \textbf{98.63}     & \textbf{90.07}      \\ \hline
\end{tabular}}
\vspace{-0.2cm}
\end{table}

\begin{table}[htbp]
\centering
\caption{$F1$ and $AUC$ (\%) of different variants for multi-level feature interaction on IID-10K dataset.}
\label{tab:5}
\setlength{\tabcolsep}{5mm}{
\begin{tabular}{r|l|ll}
\hline
Setup     & Components                             & AUC     & F1    \\ \hline
Var\#1    & w/ FPN \cite{lin2017feature}          & 93.30   & 75.10 \\
Var\#2    & w/ PANet \cite{liu2018path}           & 95.87   & 76.59 \\
Var\#3    & w/ DFPL (add)                          & 96.78    & 84.22 \\
DeFI-Net   & w/ DFPL (concat)                       & \textbf{97.24}   & \textbf{85.39} \\  \hline
\end{tabular}}
\end{table}

\noindent\textbf{Effect of REAE.} Next, we evaluated the impact of the REAE module and presented the results in Table \ref{tab:6}. Comparisons with ordinary convolutional structure Var\#1 and traditional edge extraction operator Var\#2 verify the superiority of a priori learning, significantly improving the network's performance. The selection of guidance features greatly affects the performance of the REAE. %The results of Var\#4 show that the mid-level features as guiding can give the best results, which aligns with our design above. 
Results from Var\#4 (namely our DeFI-Net model) demonstrate that using mid-level features as guidance yields the best performance. After DFPL, mid-level features fully integrate detailed and abstract information. REAE, utilizing boundary learning with strong contextual priors, effectively suppresses noisy information. %In fact, after DFPL, the mid-level features fully receive both shallow and deep information. Boundary learning with rich contextual prior guidance can effectively suppress noisy information.}

\begin{table}[tbp]
\centering
\caption{$F1$ and $AUC$ (\%) of different variants for edge supervision on IID-10K dataset.}
\label{tab:6}
\setlength{\tabcolsep}{4mm}{
\begin{tabular}{r|l|ll}
\hline
Setup     & Components                & AUC     & F1     \\ \hline
Var\#0    & w/o Edge supervision      & 98.15   & 88.14  \\
Var\#1    & w/ Conv                   & 98.35   & 89.23   \\
Var\#2    & w/ Sobel operator         & 98.38   & 89.21   \\
Var\#3    & w/ REAE (low-level guide)  & 98.42   & 89.62   \\
DeFI-Net    & w/ REAE (mid-level guide)  & \textbf{98.63}   & \textbf{90.07}   \\  \hline
\end{tabular}}
\end{table}

\noindent\textbf{Effect of SA-FF.} The fusion of high-level semantics and mid-low-level details is an effective method for enhancing localization accuracy. The SA-FF module dynamically allocates weights to different spatial positions simply. Table~\ref{tab:7} demonstrates the effectiveness of the SA-FF by comparing various fusion methods: Var\#1 (add fusion), Var\#2 (concat fusion) and Var\#3 (Dual attention~\cite{li2023edge}). The model can learn more universal fusion weights by adapting different attributes of specific inputs. Furthermore, we analyze the impact of different features used for learning weights. As indicated by Var\#4 (mid-low-level) and Var\#5 (high-level, i.e., our
DeFI-Net model), high-level features are more suitable for learning weights, probably due to their aggregation of more global information, which aids in understanding the broader context. %the high-level feature is more suitable for learning weights, probably because it aggregates more global information, which benefits understanding the global context. } 

\begin{table}[htbp]
\centering
\caption{$F1$ and $AUC$ (\%) of different variants for feature fusion on IID-10K dataset.}
\label{tab:7}
\setlength{\tabcolsep}{4 mm}{
\begin{tabular}{r|l|ll}
\hline
Setup       & Components                & AUC     & F1   \\ \hline
Var\#1    & Add fusion                & 98.38   & 89.62     \\
Var\#2    & Concat fusion             & 98.37   & 89.28     \\
Var\#3    & w/ Dual attention         & 98.13   & 90.23     \\
Var\#4    & w/ SA-FF (mid-low-level)   & 98.40   & 89.88     \\  
DeFI-Net   & w/ SA-FF (high-level)      & \textbf{98.63}     & \textbf{90.07}     \\  \hline
\end{tabular}}
\end{table}

\noindent \textbf{Effect of different loss terms.} {Furthermore, we thoroughly explored the effects of various loss terms. We have evaluated several parameter combinations for $\lambda_r$ and $\lambda_e$, corresponding to the coefficients of $L_{region}$ and $L_{edge}$, respectively. %($\lambda_r$=1, $\lambda_e$=0), ($\lambda_r$=1, $\lambda_e$=1), ($\lambda_r$=3, $\lambda_e$=1), ($\lambda_r$=1, $\lambda_e$=3), and ($\lambda_r$=5, $\lambda_e$=1), where $\lambda_r$ represents the coefficient of $L_{region}$, and $l\lambda_e$ represents the coefficient of $L_{edge}$. 
The results are shown in Table~\ref{tab:8}. When $\lambda_r$ is greater than $\lambda_e$, the model's performance improves. %However, when the weight of $L_edge$ is greater than that of $L_region$, it will harm the optimization of the model. 
We ultimately select ($\lambda_r$=5, $\lambda_e$=1) as the default parameters for the loss function.}

\begin{table}[htbp]
\centering
\caption{Ablation study with different loss weights of the loss function on IID-10K dataset.}
\label{tab:8}
\setlength{\tabcolsep}{6mm}{
\begin{tabular}{r|l|ll}
\hline
Setup      & Hyperparameters                & AUC     & F1          \\ \hline
Loss\#0    & $\lambda_r=1, \lambda_e=0$     & 98.25   & 88.87       \\
Loss\#1    & $\lambda_r=1, \lambda_e=1$     & 98.45   & 89.72       \\
Loss\#2    & $\lambda_r=3, \lambda_e=1$     & 98.30   & 89.80       \\
Loss\#3    & $\lambda_r=1, \lambda_e=3$     & 98.27   & 88.09       \\  
DeFI-Net   & $\lambda_r=5, \lambda_e=1$     & \textbf{98.63}   & \textbf{90.07}            \\  \hline
\end{tabular}}
\end{table}

\subsection{Failure Cases}
Fig. \ref{fig:lim} illustrates cases where our DeFI-Net does not yield accurate localization, particularly in scenarios with multiple tiny regions (the first two columns), dark lighting conditions (the third column), and full-size synthesis (the last column). Notably, even under these challenging conditions, DeFI-Net still demonstrates highly competitive performance compared to existing methods.

\def\subFigSz{0.22\linewidth}
\def\subFigH{0.19\linewidth}
% \captionsetup{type=figure} 
% \includegraphics[]{
\begin{figure}[t]
\centering
\vspace{-0.2cm}
\begin{minipage}[c]{0.04\linewidth}
    \caption*{\rotatebox{0}{\scriptsize{Image}}}
\end{minipage}\hfill
\begin{minipage}[r]{0.95\linewidth} \hspace{0.12cm}
    \includegraphics[width=\subFigSz, height=\subFigH]{figures/limitation/17_000000571737_ori.png}
    \includegraphics[width=\subFigSz, height=\subFigH]{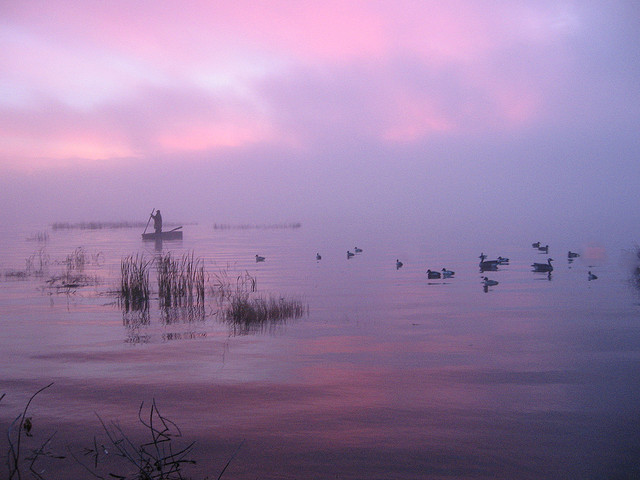}
    \includegraphics[width=\subFigSz, height=\subFigH]{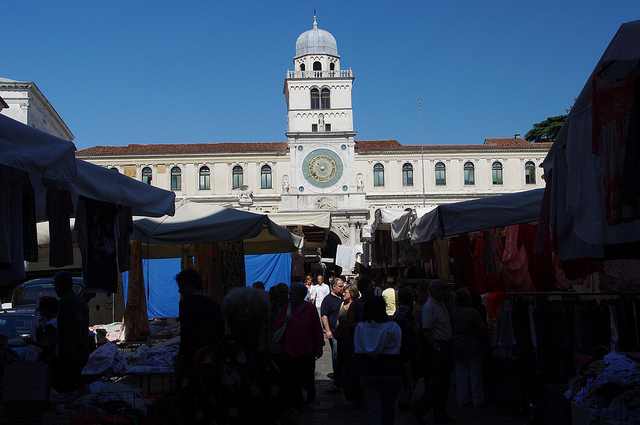}
    \includegraphics[width=\subFigSz, height=\subFigH]{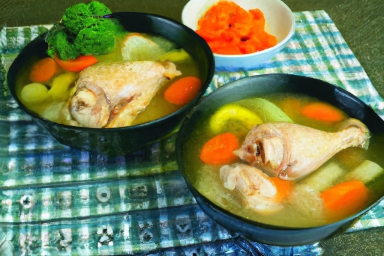}
\end{minipage}
\vspace{0.1cm}

\begin{minipage}[c]{0.04\linewidth}
    \caption*{\rotatebox{0}{\scriptsize{\;\;GT}}}
\end{minipage}\hfill
\begin{minipage}[r]{0.95\linewidth} \hspace{0.12cm}
    \includegraphics[width=\subFigSz, height=\subFigH]{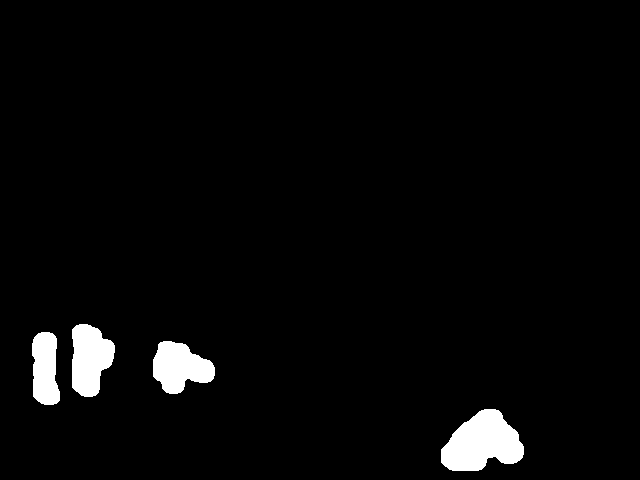}
    \includegraphics[width=\subFigSz, height=\subFigH]{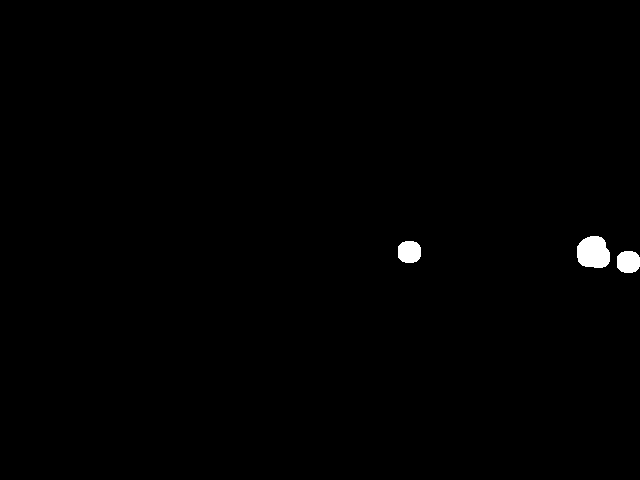}
    \includegraphics[width=\subFigSz, height=\subFigH]{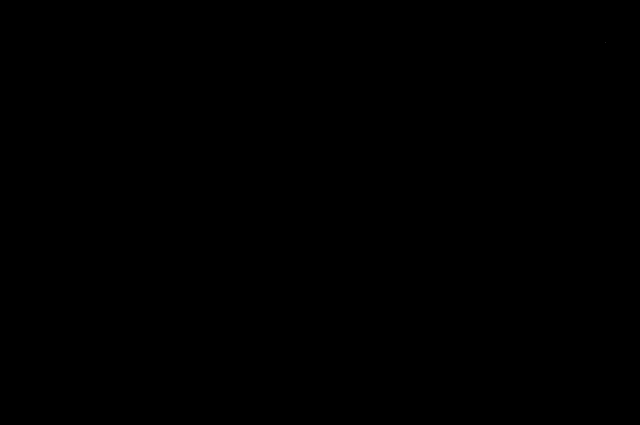}
    \includegraphics[width=\subFigSz, height=\subFigH]{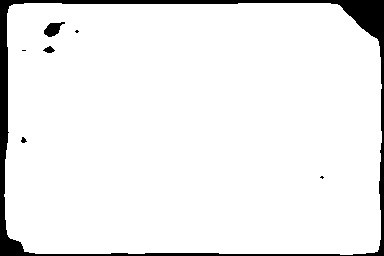}
\end{minipage}
\vspace{0.1cm}

\begin{minipage}[c]{0.04\linewidth}
    \caption*{\rotatebox{0}{\scriptsize{\,\,Ours}}}
\end{minipage}\hfill
\begin{minipage}[r]{0.95\linewidth} \hspace{0.12cm}
    \includegraphics[width=\subFigSz, height=\subFigH]{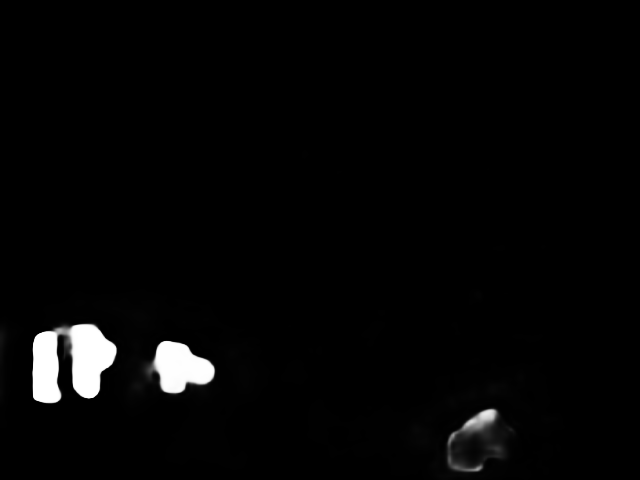}
    \includegraphics[width=\subFigSz, height=\subFigH]{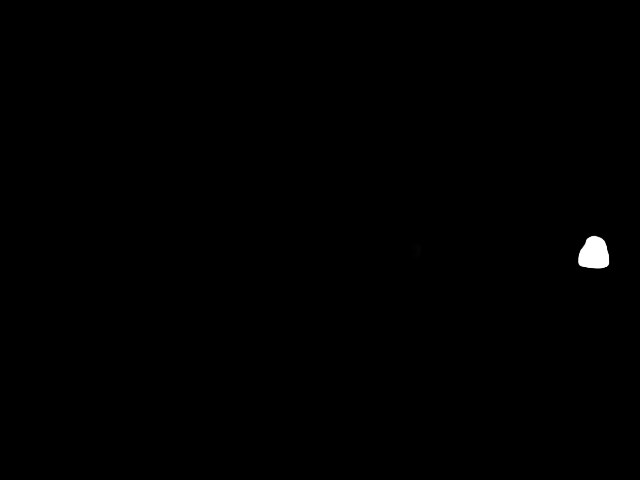}
    \includegraphics[width=\subFigSz, height=\subFigH]{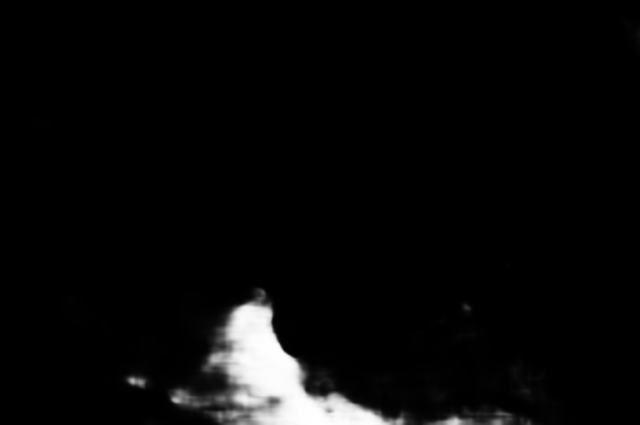}
    \includegraphics[width=\subFigSz, height=\subFigH]{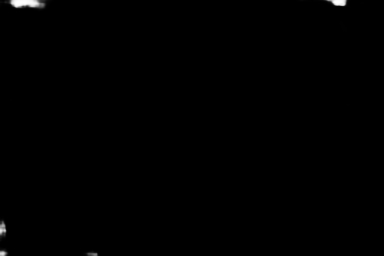}
\end{minipage}
\begin{minipage}[r]{0.95\linewidth}
\captionsetup{justification=centering}
\begin{minipage}[c]{\subFigSz}
    \caption*{\scriptsize{\quad\quad\quad(a)}}
\end{minipage}
% \hspace{-0.01in}
\begin{minipage}[c]{\subFigSz}
    \caption*{\scriptsize{\quad\quad\quad\;(b)}}
\end{minipage}
\begin{minipage}[c]{\subFigSz}
    \caption*{\scriptsize{\quad\quad\quad\;(c)}}
\end{minipage}
\begin{minipage}[c]{\subFigSz}
    \caption*{\scriptsize{\quad\quad\quad\;(d)}}
\end{minipage}
\end{minipage}
\vspace{-0.3cm}
\caption{Failure cases of the DeFI-Net, including (a) and (b) sourced from DEFACTO dataset \cite{mahfoudi2019defacto}, featuring with multiple small regions, (c) from real MSCOCO dataset \cite{lin2014microsoft} in dark lighting conditions, and (d) from the AutoSplice dataset \cite{jia2023autosplice}.}
\label{fig:lim}
\vspace{-0.2cm}
\end{figure}

\section{Conclusion}

In this work, we describe a new dense feature interaction network for image inpainting localization. Unlike existing methods, our model effectively leverages the diverse and complementary information extracted from features across various stages of the backbone architecture. We employ a dense feature pyramid structure to capture multi-level and interactive traces left by inpainting. To mitigate the impact of semantic noise on artifact representation learning, we propose a priori-guided edge enhancement strategy. Additionally, we present a spatially adaptive fusion technique to combine complementary representations, facilitating the adaptive fusion of mid-low-level features containing edge and shape information with high-level semantic features for precise localization. Extensive evaluations demonstrate the superior performance of our method.  %The results show that our performance surpasses existing detection methods and has good generalization, maintaining competitive performance even when faced with unseen inpainting methods.

Our approach has limitations when handling large inpainting regions, particularly in fully manipulated images. To address these challenges, our future research will focus on three key directions. First, we propose incorporating auxiliary tasks such as inpainting area classification, method recognition, and semantic anomaly detection to emphasize the distinctive global features between small-area inpainting, large-area inpainting, fully synthetic, and authentic images, thereby enhancing model generalization. Second, as noise pattern features are limited by susceptibility to post-processing, future work should establish a more robust noise representation through specialized neural networks and large annotated datasets to improve model adaptability. Finally, while pretrained vision models capture rich visual features and hold potential for transfer learning in inpainting detection, further research is needed to optimize architectures and training strategies for practical applications.

\bibliographystyle{IEEEtran}
\bibliography{main}

%\newpage


%\vfill
%\vspace{-1cm}
\begin{IEEEbiography}[{\includegraphics[width=1in,height=1.25in,clip,keepaspectratio]{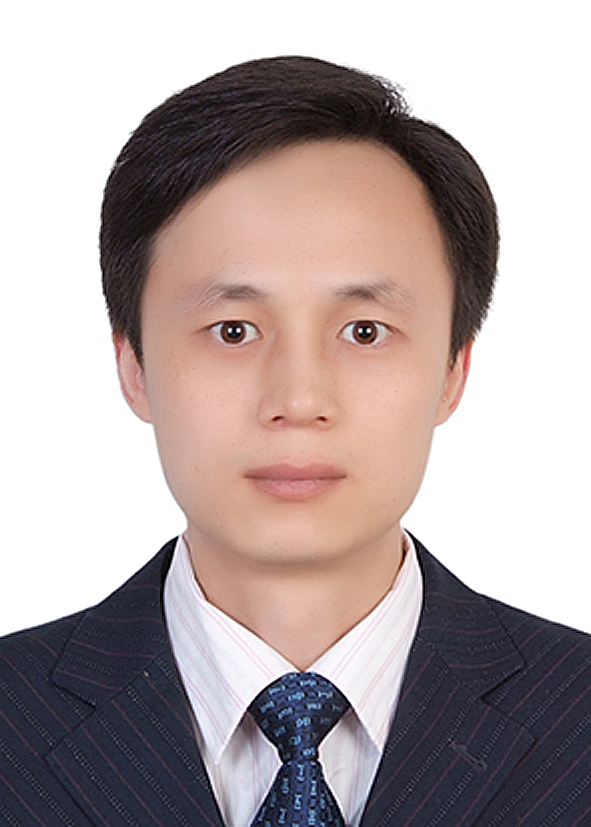}}]{Ye Yao}
received the M.S. degree in computer science and the Ph.D. degree in communication and information systems from Wuhan University, Wuhan, China, in 2005 and 2008, respectively. He was a Visiting Scholar with New Jersey Institute of Technology, Newark, NJ, USA, from December 2016 to December 2017. He is currently a professor with the School of Cyberspace, Hangzhou Dianzi University, Hangzhou. His research interests include multimedia forensics and information security.
\end{IEEEbiography}

%\vspace{-1cm}
\begin{IEEEbiography}
[{\includegraphics[width=1in,height=1.25in,clip,keepaspectratio]{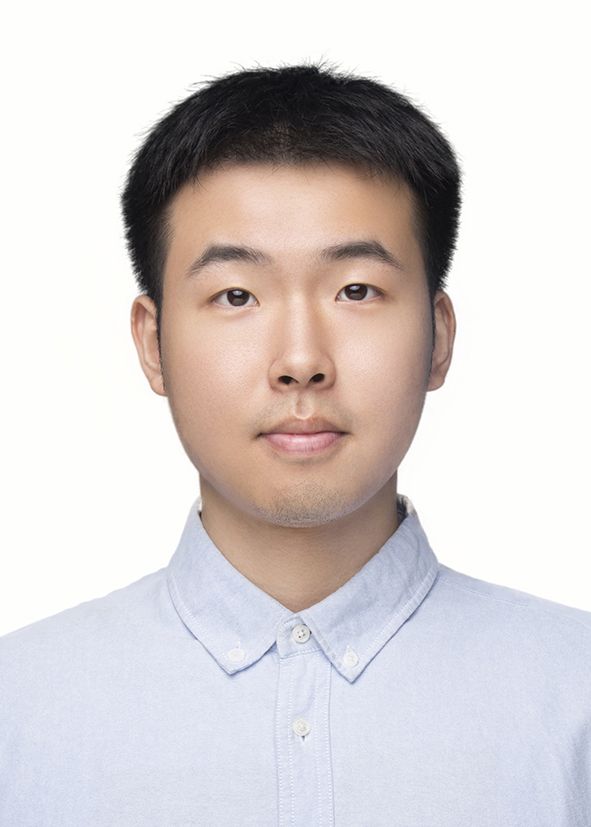}}]{Tingfeng Han}
received the B.S. degree from Hangzhou Dianzi University, Hangzhou, Zhejiang, China, in 2022. He is currently pursuing the master’s degree with the School of Cyberspace, Hangzhou Dianzi University. His current research interests include multimedia forensics and information hiding.
\end{IEEEbiography}

%\vspace{-0.5cm}
\begin{IEEEbiography}
[{\includegraphics[width=1in,height=1.25in,clip,keepaspectratio]{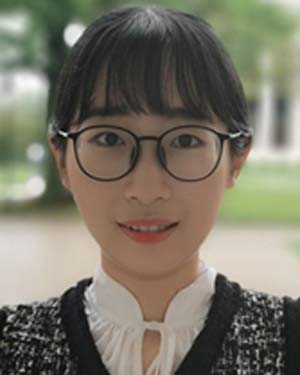}}]{Shan Jia}
received the B.S. degree in electronic and information engineering and the Ph.D. degree in communication and information systems from Wuhan University, Wuhan, China, in 2014 and 2021, respectively. She is currently a research scientist with the Department of Computer Science and Engineering, State University of New York at Buffalo, Buffalo, NY, USA. Her research interests include multimedia forensics, biometrics, and computer vision.
\end{IEEEbiography}

%\vspace{-0.5cm}
\begin{IEEEbiography}
[{\includegraphics[width=1in,height=1.25in,clip,keepaspectratio]{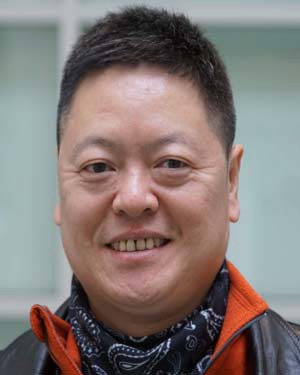}}]{Siwei Lyu}
 (Fellow, IEEE) received the M.S. and B.S. degrees in computer science and information science from Peking University, Beijing, China, in 1997 and 2000 respectively, and the Ph.D. degree in computer science from Dartmouth College, Hanover, NH, USA, in 2005. He is currently a SUNY Empire Innovation Professor with the Department of Computer Science and Engineering, University at Buffalo, State University of New York at Buffalo, Buffalo, NY, USA. His research interests include digital media forensics, computer vision, and machine learning. He is a Fellow of IEEE, IAPR and AAIA, and a Distinguished Member of ACM.
\end{IEEEbiography}

\end{document}